%% file: icml2021.tex
\documentclass{article}

\usepackage{microtype}
\usepackage{graphicx}
\usepackage{booktabs}  
\usepackage{xcolor}
\usepackage{hyperref}

\input{macro}

\usepackage[accepted]{icml2021}

\icmltitlerunning{A Policy Gradient Algorithm for Learning to Learn in Multiagent Reinforcement Learning}

\begin{document}

\twocolumn[
\icmltitle{A Policy Gradient Algorithm for Learning to Learn \\in Multiagent Reinforcement Learning}



\icmlsetsymbol{equal}{*}

\begin{icmlauthorlist}
\icmlauthor{Dong-Ki Kim}{mit,mitibm}
\icmlauthor{Miao Liu}{mitibm,ibm}
\icmlauthor{Matthew Riemer}{mitibm,ibm}
\icmlauthor{Chuangchuang Sun}{mit,mitibm}
\icmlauthor{Marwa Abdulhai}{mit,mitibm}
\\
\icmlauthor{Golnaz Habibi}{mit,mitibm}
\icmlauthor{Sebastian Lopez-Cot}{mit,mitibm}
\icmlauthor{Gerald Tesauro}{mitibm,ibm}
\icmlauthor{Jonathan P. How}{mit,mitibm}
\end{icmlauthorlist}

\icmlaffiliation{mit}{MIT-LIDS}
\icmlaffiliation{ibm}{IBM-Research}
\icmlaffiliation{mitibm}{MIT-IBM Watson AI Lab}

\icmlcorrespondingauthor{Dong-Ki Kim}{dkkim93@mit.edu}
\icmlcorrespondingauthor{Miao Liu}{miao.liu1@ibm.com}

\icmlkeywords{Multiagent reinforcement learning, Meta-learning, Non-stationarity}

\vskip 0.3in
]



\printAffiliationsAndNotice{}  

\setlength{\abovedisplayskip}{3.5pt}
\setlength{\belowdisplayskip}{3.5pt}
\setlength{\abovedisplayshortskip}{3.5pt}
\setlength{\belowdisplayshortskip}{3.5pt}

\begin{abstract}
A fundamental challenge in multiagent reinforcement learning is to learn beneficial behaviors in a shared environment with other simultaneously learning agents. In particular, each agent perceives the environment as effectively non-stationary due to the changing policies of other agents. Moreover, each agent is itself constantly learning, leading to natural non-stationarity in the distribution of experiences encountered. In this paper, we propose a novel meta-multiagent policy gradient theorem that directly accounts for the non-stationary policy dynamics inherent to multiagent learning settings. This is achieved by modeling our gradient updates to consider both an agent’s own non-stationary policy dynamics and the non-stationary policy dynamics of other agents in the environment. We show that our theoretically grounded approach provides a general solution to the multiagent learning problem, which inherently comprises all key aspects of previous state of the art approaches on this topic. We test our method on a diverse suite of multiagent benchmarks and demonstrate a more efficient ability to adapt to new agents as they learn than baseline methods across the full spectrum of mixed incentive, competitive, and cooperative domains.
\end{abstract}

\input{introduction}
\input{problem_statement}
\input{approach}
\input{related_work}
\input{experiment}
\input{conclusion}

\section*{Acknowledgements}
Research funded by IBM, Samsung (as part of the MIT-IBM Watson AI Lab initiative) and computational support through Amazon Web Services. Dong-Ki Kim was also supported by a Kwanjeong Educational Foundation Fellowship. The authors would like to thank anonymous reviewers for their helpful comments.

\bibliography{icml2021}
\bibliographystyle{icml2021}

\newpage
\appendix
\input{appendix}
\end{document}

%% file: macro.tex
\usepackage{amsmath}
\usepackage{color}
\usepackage{soul}
\usepackage{xspace}
\usepackage[capitalize]{cleveref}
\usepackage{dsfont}
\usepackage{bm}
\usepackage{bbm}
\usepackage{nicefrac}  
\usepackage{amsthm}
\usepackage{amsfonts}
\usepackage{amssymb}
\usepackage{array}
\usepackage{mathtools}
\usepackage{caption}
\usepackage{subcaption}
\usepackage{fontawesome}
\usepackage{thmtools}
\usepackage{thm-restate}
\usepackage{lipsum}
\usepackage{pifont}
\usepackage{makecell}
\usepackage[tight-spacing=true]{siunitx}
\usepackage{multirow}
\usepackage{afterpage}
\usepackage{tabularx} 
\usepackage{url}
\usepackage{enumitem}  
\usepackage{wrapfig}  
\usepackage{textcomp}

\Crefname{figure}{Figure}{Figures}
\crefname{figure}{Figure}{Figures}
\crefname{table}{Table}{Tables}
\crefformat{table}{Table~#1}
\crefformat{equation}{Equation (#1)}
\Crefformat{equation}{Equation (#1)}
\crefformat{algorithm}{Algorithm~#1}
\crefformat{appendix}{Appendix~#1}
\Crefformat{appendix}{Appendix~#1}
\crefformat{appendices}{Appendices~#1}
\Crefformat{appendices}{Appendices~#1}

\newcommand{\header}[1]{\textbf{#1}\,\,}

\newcommand\sdots{\hbox to 1em{.\hss.\hss.}} 
\DeclareMathAlphabet\mathbfcal{OMS}{cmsy}{b}{n}  
\newcommand{\smallsum}{\textstyle\sum\limits}

\newtheorem{question}{Question}

\newif\ifcomments
\commentsfalse

\ifcomments
	\newcommand{\dXX}[1]{\color{red}DK: (#1)\color{black}\xspace}  
	\newcommand{\mlXX}[1]{\color{cyan}ML: (#1)\color{black}\xspace}  
	\newcommand{\mrXX}[1]{\color{blue}MR: (#1)\color{black}}  
	\newcommand{\maXX}[1]{\color{red}MA: (#1)\color{black}\xspace}  
	\newcommand{\cXX}[1]{\color{red}CS: (#1)\color{black}\xspace}  
	\newcommand{\ghXX}[1]{\color{olive}GH: (#1)\color{black}}  
	\newcommand{\slXX}[1]{\color{magenta}SL: (#1)\color{black}}  
	\newcommand{\gtXX}[1]{\color{purple}GT: (#1)\color{black}\xspace}  
	\newcommand{\jXX}[1]{\color{orange}JH: (#1)\color{black}\xspace}  
\else
    \newcommand{\dXX}[1]{}  
	\newcommand{\mlXX}[1]{}  
	\newcommand{\mrXX}[1]{}  
	\newcommand{\cXX}[1]{}  
	\newcommand{\maXX}[1]{}  
	\newcommand{\ghXX}[1]{}  
	\newcommand{\slXX}[1]{}  
	\newcommand{\gtXX}[1]{}  
	\newcommand{\jXX}[1]{}  
\fi

\newcommand*{\QEDB}{\null\nobreak\hfill\ensuremath{\square}}%

%% file: introduction.tex
\section{Introduction}\label{sec:introduction}

Learning in multiagent settings is inherently more difficult than single-agent learning because an agent interacts both with the environment and other agents~\citep{Busoniu2010}.
Specifically, the fundamental challenge in multiagent reinforcement learning (MARL) is the difficulty of learning optimal policies 
in the presence of other simultaneously learning agents because their changing behaviors jointly affect the environment's transition and reward function. 
This dependence on non-stationary policies renders the Markov property invalid from the perspective of each agent, requiring agents to adapt their behaviors with respect to potentially large, unpredictable, and endless changes in the policies of fellow agents~\citep{papoudakis19nonstationarity}.
In such environments, it is also critical that agents adapt to the changing behaviors of others in a sample-efficient manner as it is likely that their policies could update again after a small number of interactions.
Therefore, effective agents should consider the learning of other agents and adapt quickly to their non-stationary behaviors. 
Otherwise, undesirable outcomes may arise when an agent is constantly lagging in its ability to deal with the current policies of other agents. 

Our paper proposes a new framework based on meta-learning to address the inherent non-stationarity of MARL.
Meta-learning (also referred to as learning to learn) was recently shown to be a promising methodology for fast adaptation in multiagent settings. 
The framework by~\citet{alshedivat2018continuous}, for example, introduces a meta-optimization scheme in which a meta-agent can adapt more efficiently to changes in a new opponent's policy after collecting only a handful of interactions.
The key idea is to model the meta-agent's own learning process so that its updated policy performs better than an evolving opponent.
However, prior work does not directly consider the learning processes of other agents during the meta-optimization process, instead treating the other agents as external factors and assuming the meta-agent cannot influence their future policies.
As a result, prior work on multiagent meta-learning fails to consider an important property of MARL: other agents in the environment are also learning and adapting their own policies based on interactions with the meta-agent. 
Thus, the meta-agent is missing an opportunity to influence the others' future policies through these interactions, which could be used to improve its own performance.
  
\noindent\header{Our contributions.} With this insight, we make the following primary contributions in this paper:
\begin{enumerate}[leftmargin=*, wide, labelindent=0pt, topsep=0pt, label=\arabic*)]
    \itemsep0em 
    \item \textbf{New theorem:} We derive a new \textit{meta-multiagent policy gradient theorem (Meta-MAPG)} that, for the first time, directly models the learning processes of all agents in the environment within a single objective function. We achieve this by extending past work that developed the meta-policy gradient theorem (Meta-PG) in the context of a single agent multi-task RL setting~\citep{alshedivat2018continuous} to the full generality of a multiagent game where every agent learns with a Markovian update function.
    \item \textbf{Theoretical analysis and comparisons:} Our analysis of Meta-MAPG reveals an inherently added term, not present in~\citet{alshedivat2018continuous}, closely related to the process of shaping the other agents' learning dynamics in the framework of~\citet{foerster17lola}. 
    As such, our work 
    contributes a theoretically grounded framework that unifies the collective benefits of previous work by~\citet{alshedivat2018continuous} and~\citet{foerster17lola}. 
    Moreover, we formally demonstrate that Meta-PG can be considered a special case of Meta-MAPG if we assume the learning of the other agents in the environment is constrained to be independent of the meta-agent behavior. 
    However, this limiting assumption does not typically hold in practice, and we demonstrate with a motivating example that it can lead to a total divergence of learning that is directly addressed by the correction we derive for the meta-gradient. 
    \item \textbf{Empirical evaluation:} We evaluate Meta-MAPG on a diverse suite of multiagent domains, including the full spectrum of mixed incentive, competitive, and cooperative environments. 
    Our experiments demonstrate that, in contrast with previous state-of-the-art approaches on this topic, Meta-MAPG consistently results in a superior ability to adapt in the presence of novel agents as they learn.
\end{enumerate}

%% file: problem_statement.tex
\section{Understanding Non-Stationarity in MARL}\label{sec:background-non-stationarity}
In MARL, when the policies of all other agents in the environment are stationary, they can be considered to be effectively a part of the environment from the perspective of each agent, resulting in a stationary single agent Markov decision process for each agent to learn from. 
However, in practice, the other agents in the environment are not fixed, but rather always learning from their recent experiences. As we detail in this section, this results in the inherent non-stationarity of MARL because Markovian update functions for each agent induce a Markov chain of joint policies. Indeed, when the policies of other agents constantly change, the problem is effectively non-stationary from each agent's perspective if the other agents are viewed as part of the environment.  

\subsection{Stochastic Games}
Interactions between multiple agents can be represented by stochastic games~\citep{shapley53stochastic}. 
Specifically, an $n$-agent stochastic game is defined as a tuple $\mathcal{M}_{n}\!=\!\langle \mathbfcal{I}, \mathbfcal{S}, \mathbfcal{A}, \mathcal{P}, \mathbfcal{R}, \gamma \rangle$;
$\mathbfcal{I}\!=\!\{1,\sdots,n\}$ is the set of $n$ agents, 
$\mathbfcal{S}$ is the set of states, 
$\mathbfcal{A}\!=\!\times_{i \in \mathcal{I}} \mathcal{A}^{i}$ is the set of action spaces, 
$\mathcal{P}\!:\!\mathbfcal{S}\times\mathbfcal{A}\mapsto \mathbfcal{S}$ is the state transition probability function,
$\mathbfcal{R}\!=\!\times_{i \in \mathcal{I}} \mathcal{R}^{i}$ is the set of reward functions, and
$\gamma\!\in\![0,1)$ is the discount factor.
We typeset sets in bold for clarity.
Each agent $i$ executes an action at each timestep $t$ according to its stochastic policy $a^i_t\!\sim\!\pi^{i}(a^{i}_{t}|s_{t},\phi^i)$ parameterized by $\phi^{i}$, where $s_{t}\!\in\!\mathbfcal{S}$.
A joint action $\bm{a}_{\bm{t}}\!=\!\{a^{i}_t,\bm{a}^{\bm{-i}}_{\bm{t}}\}$ yields a transition from the current state $s_{t}$ to the next state $s_{t+1}\!\in\!\mathbfcal{S}$ with probability $\mathcal{P}(s_{t+1}|s_t,\bm{a}_{\bm{t}})$, where the notation $\bm{-i}$ indicates all other agents with the exception of agent $i$.
Agent $i$ then obtains a reward according to its reward function $r^{i}_t\!=\!\mathcal{R}^i(s_t,\bm{a}_{\bm{t}})$.
At the end of an episode, the agents collect a trajectory $\tau_{\bm{\phi}}$ under the joint policy with parameters $\bm{\phi}$, where $\tau_{\bm{\phi}}\!:=\!(s_{0},\bm{a}_{\bm{0}},\bm{r}_{\bm{0}},\sdots,\bm{r}_{\bm{H}})$, $\bm{\phi}\!=\!\{\phi^{i},\bm{\phi}^{\bm{-i}}\}$ represents the joint parameters of all policies, $\bm{r}_{\bm{t}}\!=\!\{r^{i}_{t},\bm{r}^{\bm{-i}}_{\bm{t}}\}$ is the joint reward, and $H$ is the episode horizon.

\begin{figure}[t]
   \captionsetup[subfigure]{skip=0pt, aboveskip=2pt}
   \begin{subfigure}[b]{\linewidth}
        \centering
        \includegraphics[height=3.1cm]{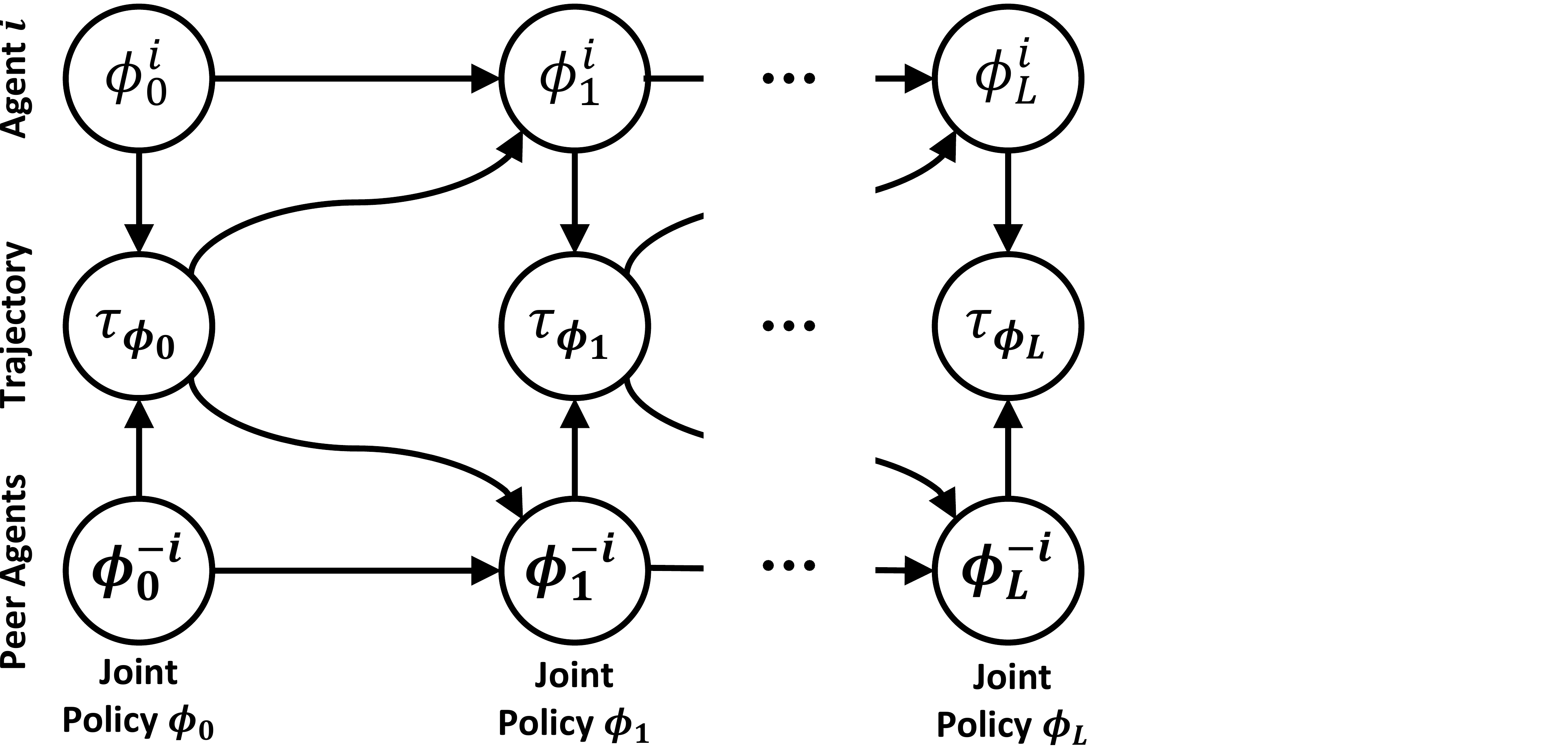}
        \caption{}
        \label{fig:markov-chain}
    \end{subfigure}
    \begin{subfigure}[b]{\linewidth}
        \centering
        \includegraphics[height=3.1cm]{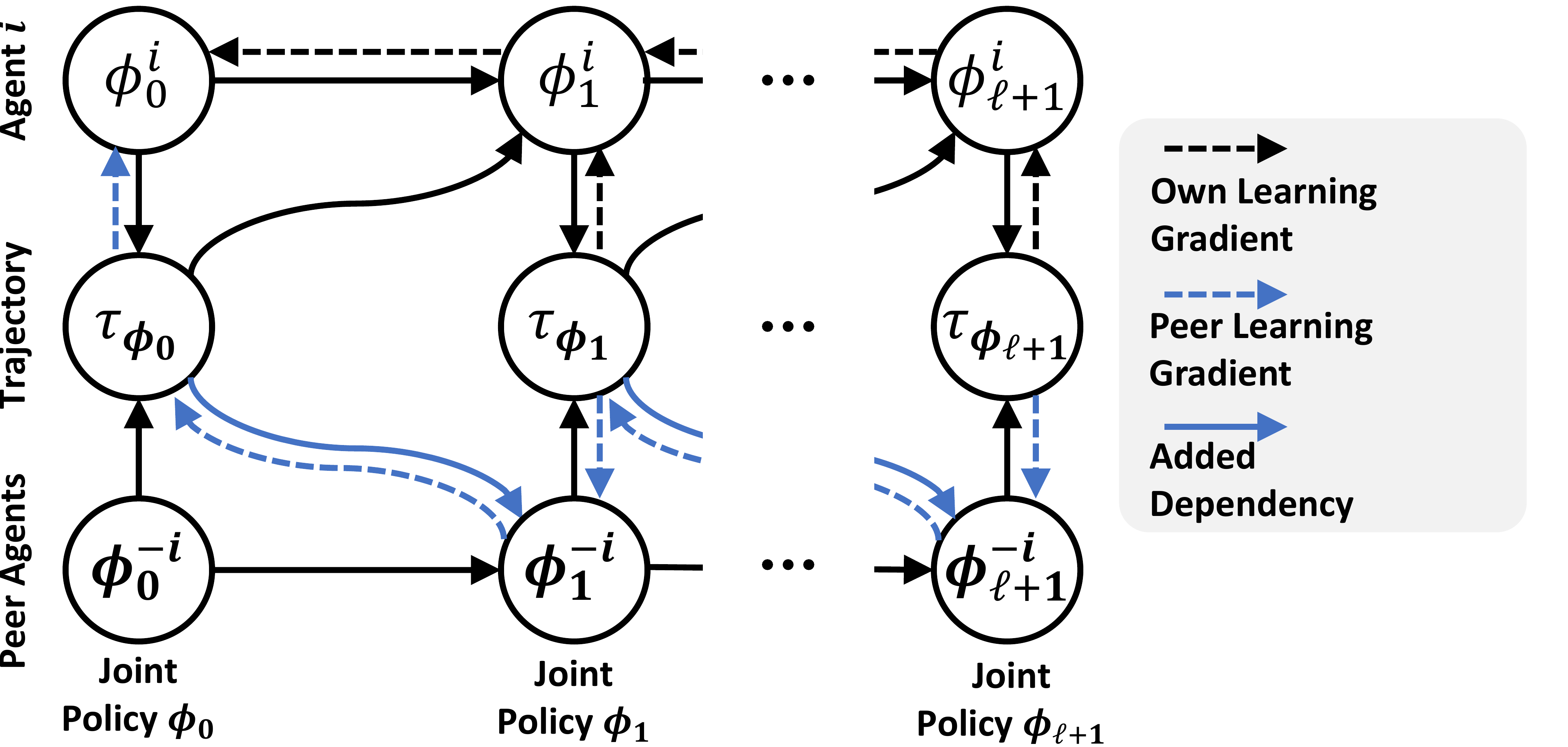}
        \caption{}
        \label{fig:graph-meta-mapg}
    \end{subfigure}
    \vspace{-0.8cm}
    \caption{
        \textbf{(a)} A Markov chain of joint policies representing the inherent non-stationarity of MARL. Each agent updates its policy leveraging a Markovian update function, resulting in a change to the joint policy. 
        \textbf{(b)} A probabilistic graph for Meta-MAPG. Unlike Meta-PG, our approach actively influences the future policies of other agents as well through the peer learning gradient.
        }
    \vskip-0.18in
\end{figure}

\subsection{A Markov Chain of Joint Policies}\label{sec:markov-chain-of-policies}
The perceived non-stationarity in multiagent settings results from a distribution of sequential joint policies, which can be represented by a Markov chain.
Formally, a Markov chain of policies begins from a stochastic game between agents with an initial set of joint policies parameterized by $\bm{\phi}_{\bm{0}}\!=\!\{\phi^{i}_{0},\bm{\phi}^{\bm{-i}}_{\bm{0}}\}$. 
We assume that each agent updates its policy leveraging a Markovian update function $\phi^{i}_{1}\!\sim\!U^{i}(\tau_{\bm{\phi}_{\bm{0}}},\phi^{i}_{0})$ that changes the policy after every $K$ trajectories. 
After this time period, each agent $i$ updates its policy in order to maximize the expected return expressed as its value function:
\begin{gather}
\!\!\!V^{i}_{\!\bm{\phi}_{\bm{0}}}\!(s_{0})\!=\!\mathbb{E}_{p(\tau_{\bm{\phi}_{\bm{0}}}\!|\bm{\phi}_{\bm{0}})}\!\Big[\!\smallsum_{t=0}^{H}\!\!\gamma^{t}r^{i}_{t}|s_{0}\hspace{-0.1em}\Big]\!\!=\!\mathbb{E}_{p(\tau_{\bm{\phi}_{\bm{0}}}\!|\bm{\phi}_{\bm{0}})}\!\Big[\!G^{i}(\tau_{\!\bm{\phi}_{\bm{0}}})\!\Big]\!,\label{eqn:value-function}
\end{gather}
where $G^{i}$ denotes agent $i$'s discounted return from the beginning of an episode with initial state $s_0$.
The joint policy update results in a transition from $\bm{\phi}_{\bm{0}}$ to the updated set of joint parameters $\bm{\phi}_{\bm{1}}$.
The Markov chain continues for a maximum chain length of $L$ (see~\cref{fig:markov-chain}). 
This Markov chain perspective highlights the following inherent aspects of the experienced non-stationarity:

\begin{enumerate}[leftmargin=*, wide, labelindent=0pt, topsep=0pt, label=\arabic*)]
    \itemsep0em 
    \item \textbf{Sequential dependency:} The future joint policy parameters $\bm{\phi}_{\bm{1:L}}\!=\!\{\bm{\phi}_{\bm{1}}, \sdots,\bm{\phi}_{\bm{L}}\}$ sequentially depend on $\bm{\phi}_{\bm{0}}$ since a change in $\tau_{\bm{\phi}_{\bm{0}}}$ results in a change in $\bm{\phi}_{\bm{1}}$, which in turn affects $\tau_{\bm{\phi}_{\bm{1}}}$ and all successive joint policy updates up to $\bm{\phi}_{\bm{L}}$. 
    \item\textbf{Controllable levels of non-stationarity:} As in~\citet{alshedivat2018continuous} and~\citet{foerster17lola}, we assume stationary policies during the collection of $K$ trajectories, and that the joint policy update happens afterward. In such a setting, it is possible to control the non-stationarity by adjusting the $K$ and $H$ hyperparameters: smaller $K$ and $H$ increase the rate that agents change their policies, leading to a higher degree of non-stationarity in the environment.
\end{enumerate}

%% file: approach.tex
\section{Learning to Learn in MARL}\label{sec:approach}   
This section explores learning policies that can adapt quickly to non-stationarity in the policies of other agents in the environment.      
To achieve this, we leverage meta-learning and devise a new \textit{meta-multiagent policy gradient theorem} that exploits the inherent sequential dependencies of MARL discussed in the previous section. 
Specifically, our meta-agent addresses this non-stationarity by considering its current policy's impact on its own adapted policies while actively influencing the future policies of other agents as well by inducing changes to the distribution of trajectories.
In this section, we first outline the meta-optimization objective of MARL and then derive our policy gradient theorem to optimize this objective. 
Finally, we discuss how to interpret our meta-policy gradient theorem with respect to prior work. 

\subsection{Gradient Based Meta-Optimization in MARL}
We formalize the meta-objective of MARL as optimizing meta-agent $i$'s initial policy parameters $\phi^{i}_{0}$ so that it maximizes the expected adaptation performance over a Markov chain of policies drawn from a stationary initial distribution of policies for the other agents $p(\bm{\phi}^{\bm{-i}}_{\bm{0}})$:
\begin{gather}
\max_{\phi^{i}_{0}}\mathbb{E}_{p(\bm{\phi}^{\bm{-i}}_{\bm{0}})}\Big[\smallsum_{\ell=0}^{L-1}V^{i}_{\bm{\phi}_{\bm{0}:\bm{\ell+1}}}(s_0,\phi^{i}_{0})\Big],\label{eqn:meta-value-function}\\
V^{i}_{\bm{\phi}_{\bm{0:\ell+1}}}\!(s_0,\!\phi^{i}_{0})\!=\!\mathbb{E}_{p(\tau_{\bm{\phi}_{\bm{0:\ell}}}\!|\bm{\phi}_{\bm{0:\ell}})}\!\Big[\mathbb{E}_{p(\tau_{\bm{\phi}_{\bm{\ell+1}}}\!|\bm{\phi}_{\bm{\ell+1}})}\!\big[G^{i}\!(\tau_{\!\bm{\phi}_{\bm{\ell+1}}})\big]\!\Big]\nonumber\!,
\end{gather}
where $\tau_{\bm{\phi}_{\bm{0:\ell}}}\!=\!\{\tau_{\bm{\phi}_{\bm{0}}},\sdots,\tau_{\bm{\phi}_{\bm{\ell}}}\}$ and $V^{i}_{\bm{\phi}_{\bm{0:\ell+1}}}(s_0,\phi^{i}_{0})$ denotes the meta-value function.
This meta-value function generalizes the notion of each agent's primitive value function for the current set of policies $V^{i}_{\bm{\phi}_{\bm{0}}}(s_{0})$ over the length of the Markov chain of policies. 
In this work, we assume that the Markov chain of policies is governed by a policy gradient update function that corresponds to what is generally referred to as the inner-loop optimization in the literature:
\begin{align}\label{eqn:inner-loop}
\begin{split}
\phi^{i}_{\ell+1}\!=& U^{i}(\tau_{\bm{\phi}_{\bm{\ell}}},\phi^{i}_{\ell})\!:=\!\phi^{i}_{\ell}\!+\!\alpha^i\nabla_{\!\phi^{i}_{\ell}}\mathbb{E}_{p(\tau_{\bm{\phi}_{\bm{\ell}}}|\bm{\phi}_{\bm{\ell}})}\Big[G^{i}(\tau_{\bm{\phi}_{\bm{\ell}}})\Big]\!,\\
\bm{\phi}^{\bm{-i}}_{\!\bm{\ell+1}}\!\!=& U^{\!-i}\!(\tau_{\!\bm{\phi}_{\bm{\ell}}},\!\phi^{\!-i}_{\ell})\!\!:=\!\bm{\phi}^{\!\bm{-i}}_{\bm{\ell}}\!\!+\!\hspace{-0.1em}\bm{\alpha}^{\!\bm{-i}}\nabla_{\!\!\!\bm{\phi}^{\!\bm{-i}}_{\bm{\ell}}}\mathbb{E}_{p(\tau_{\bm{\phi}_{\bm{\ell}}}\!|\bm{\phi}_{\bm{\ell}})}\!\hspace{-0.1em}\Big[\!\bm{G}^{\bm{-i}}\!(\tau_{\bm{\phi}_{\bm{\ell}}})\!\Big]\!,
\raisetag{38pt}
\end{split}
\end{align}
where $\alpha^i$ and $\bm{\alpha}^{\bm{-i}}$ denote the learning rates.

\subsection{The Meta-Multiagent Policy Gradient Theorem}\label{sec:meta-mapg-theorem}
A meta-agent needs to account for both its own learning process and the learning processes of other peer agents in the environment to fully address the inherent non-stationarity of MARL. 
We will now demonstrate that our generalized meta-policy gradient includes terms that explicitly account for the effect a meta-agent's current policy will have on its own adapted future policies as well as the future policies of peers interacting with it.

\textbf{Theorem 1.} (Meta-Multiagent Policy Gradient Theorem) \textit{For any stochastic game $\mathcal{M}_{n}$, the gradient of the meta-value function for agent $i$ at state $s_0$ with respect to current policy parameters $\phi_0^i$ evolving in the environment along with other peer agents using initial parameters $\bm{\phi}_{\bm{0}}^{\bm{-i}}$ is:}
\begin{align}\label{eqn:meta-multiagent-policy-gradient}
\begin{split}
&\nabla_{\!\!\phi^{i}_{0}}\!V^{i}_{\!\bm{\phi}_{\bm{0:\ell+1}}}\!(s_0,\!\phi^{i}_{0})\!=\!\mathbb{E}_{p(\tau_{\bm{\phi}_{\bm{0:\ell}}}\!|\bm{\phi}_{\bm{0:\ell}})}\!\Big[\!\mathbb{E}_{p(\tau_{\bm{\phi}_{\bm{\ell+1}}}\!|\bm{\phi}_{\bm{\ell+1}})}\!\big[\!G^{i}(\hspace{-0.1em}\tau_{\hspace{-0.1em}\bm{\phi}_{\bm{\ell+1}}}\hspace{-0.1em})\\
&\;\;\big(\underbrace{\nabla_{\!\!\phi^{i}_{0}}\!\log\!\pi(\tau_{\bm{\phi}_{\bm{0}}}\!|\phi^{i}_{0})}_{\text{Current Policy}}\!+\!\underbrace{\smallsum\nolimits_{\ell'\!=0}^{\ell}\!\!\nabla_{\!\!\phi^{i}_{0}}\!\log\!\pi(\tau_{\bm{\phi}_{\bm{\ell'+1}}}\!|\phi^{i}_{\ell'+1})}_{\text{Own Learning}}+\\
&\;\;\;\underbrace{\smallsum\nolimits_{\ell'\!=0}^{\ell}\!\!\nabla_{\!\!\phi^{i}_{0}}\!\log\!\pi(\tau_{\bm{\phi}_{\bm{\ell'+1}}}\!|\bm{\phi}^{\bm{-i}}_{\bm{\ell'+1}})}_{\text{Peer Learning}}\big)\big]\Big].
\raisetag{63pt}
\end{split}
\end{align}
\textit{Proof.} \hspace{0.2em}See~\cref{sec:proof-meta-multiagent-pg} for a detailed proof.\QEDB

In particular, Meta-MAPG has three primary terms. The first term corresponds to the standard policy gradient with respect to the current policy parameters used during the initial trajectory. Meanwhile, the second term $\nabla_{\phi^{i}_{0}}\log\pi(\tau_{\bm{\phi}_{\bm{\ell'+1}}}|\phi^{i}_{\ell'+1})$ explicitly differentiates through $\log\pi(\tau_{\bm{\phi}_{\bm{\ell'+1}}}|\phi^{i}_{\ell'+1})$ with respect to $\phi^{i}_{0}.$ 
This enables a meta-agent to model its own learning dynamics and account for the impact of $\phi^{i}_{0}$ on its eventual adapted parameters $\phi^{i}_{\ell'+1}$. 
By contrast, the last term $\nabla_{\phi^{i}_{0}}\log\pi(\tau_{\bm{\phi}_{\bm{\ell'+1}}}|\bm{\phi}^{\bm{-i}}_{\bm{\ell'+1}})$ aims to additionally compute gradients through the sequential dependency between the meta-agent's initial policy $\phi^{i}_{0}$ and the future policies of other agents in the environment $\bm{\phi}^{\bm{-i}}_{\bm{1:\ell+1}}$.
As a result, the peer learning term enables a meta-agent to learn to change $\tau_{\bm{\phi}_{\bm{0}}}$ in a way that influences the future policies of other agents and improves its adaptation performance over the Markov chain of policies. 

Interestingly, the peer learning term that naturally arises when taking the gradient in Meta-MAPG is similar to a term previously considered in the literature by \citet{foerster17lola}. 
However, for the Learning with Opponent Learning Awareness (LOLA) approach \citep{foerster17lola}, this term was derived in an alternate way following a first order Taylor approximation with respect to the value function. Moreover, the own learning term previously appeared in the derivation of Meta-PG \citep{alshedivat2018continuous}.
Indeed, it is quite surprising to see how taking a principled policy gradient while leveraging a more general set of assumptions leads to a unification of the benefits of key prior work \citep{alshedivat2018continuous,foerster17lola} on adjusting to the learning behavior of other agents in MARL. 

\subsection{Connection to the Meta-Policy Gradient Theorem}\label{sec:meta-pg-theorem}
The framework of~\citet{alshedivat2018continuous} derived the meta-policy gradient theorem for optimizing a setup like this. 
However, it is important to note that they derived this gradient while making the implicit assumption to ignore the sequential dependence of the future parameters of other agents on $\phi^{i}_{0}$, resulting in the missing peer learning term:
\begin{align}\label{eqn:meta-policy-gradient}
\begin{split}
&\nabla_{\!\!\phi^{i}_{0}}\!V^{i}_{\!\bm{\phi}_{\bm{0:\ell+1}}}\!(s_0,\!\phi^{i}_{0})\!=\!\mathbb{E}_{p(\tau_{\bm{\phi}_{\bm{0:\ell}}}\!|\bm{\phi}_{\bm{0:\ell}})}\!\Big[\!\mathbb{E}_{p(\tau_{\bm{\phi}_{\bm{\ell+1}}}\!|\bm{\phi}_{\bm{\ell+1}})}\!\big[\!G^{i}(\hspace{-0.1em}\tau_{\hspace{-0.1em}\bm{\phi}_{\bm{\ell+1}}}\hspace{-0.1em})\\
&\;\;\big(\underbrace{\nabla_{\!\!\phi^{i}_{0}}\!\log\!\pi(\tau_{\bm{\phi}_{\bm{0}}}\!|\phi^{i}_{0})}_{\text{Current Policy}}\!+\!\underbrace{\smallsum\nolimits_{\ell'\!=0}^{\ell}\!\!\nabla_{\!\!\phi^{i}_{0}}\!\log\!\pi(\tau_{\bm{\phi}_{\bm{\ell'+1}}}\!|\phi^{i}_{\ell'+1})}_{\text{Own Learning}}\big)\big]\Big]\!.
\raisetag{30pt}
\end{split}
\end{align}
\textbf{Remark 1.} \textit{
Meta-PG can be considered as a special case of Meta-MAPG when assuming that other agents' learning in the environment is independent of the meta-agent's behavior.}

\textit{Proof.} \hspace{0.2em}See~\cref{sec:proof-meta-pg} for a corollary.\QEDB

Specifically, Meta-PG treats other agents as if they are external factors whose learning it cannot affect.
For example, a meta-agent in~\citet{alshedivat2018continuous} competes against an opponent that has been pre-trained with self-play. The next agent after the current interaction is then loaded as the same opponent with one more step of training with self-play. 
As a result of this contrived setup, the opponent's policy update is based on data collected without a meta-agent presented in the environment and Meta-PG can get away with assuming that the peer's learning process is not a function of a meta-agent's behavior.
However, we note that this assumption does not hold in practice for the general multi-agent learning settings we explore in this work. 
As highlighted in~\cref{sec:background-non-stationarity}, every agent's policy jointly affects the environment's transition and reward functions, resulting in the inherent sequential dependencies of MARL. Therefore, a meta-agent's behavior can affect the future policies of its peers, and these cannot be considered as external factors.

\noindent\header{Probabilistic model perspective.}
Probabilistic models for Meta-PG and Meta-MAPG are depicted in~\cref{fig:graph-meta-mapg}.
As shown by the own learning gradient direction, a meta-agent $i$ optimizes $\phi^{i}_{0}$ by accounting for the impact of $\phi^{i}_{0}$ on its updated parameters $\bm{\phi}^{\bm{i}}_{\bm{1:\ell+1}}$ and adaptation performance $G^{i}(\tau_{\bm{\phi}_{\bm{\ell+1}}})$.
However,  Meta-PG considers the other agents as external factors that cannot be influenced by the meta-agent, as indicated by the absence of the sequential dependence between $\tau_{\bm{\phi}_{\bm{0:\ell}}}$ and $\bm{\phi}^{\bm{-i}}_{\bm{1:\ell+1}}$ in~\Cref{fig:graph-meta-mapg}. 
As a result, the meta-agent loses an opportunity to influence the future policies of other agents and further improve its adaptation performance, which can cause undesirable learning failures as highlighted by the following motivating example.

\begin{figure}[t]
    \centering
    \includegraphics[height=2.9cm]{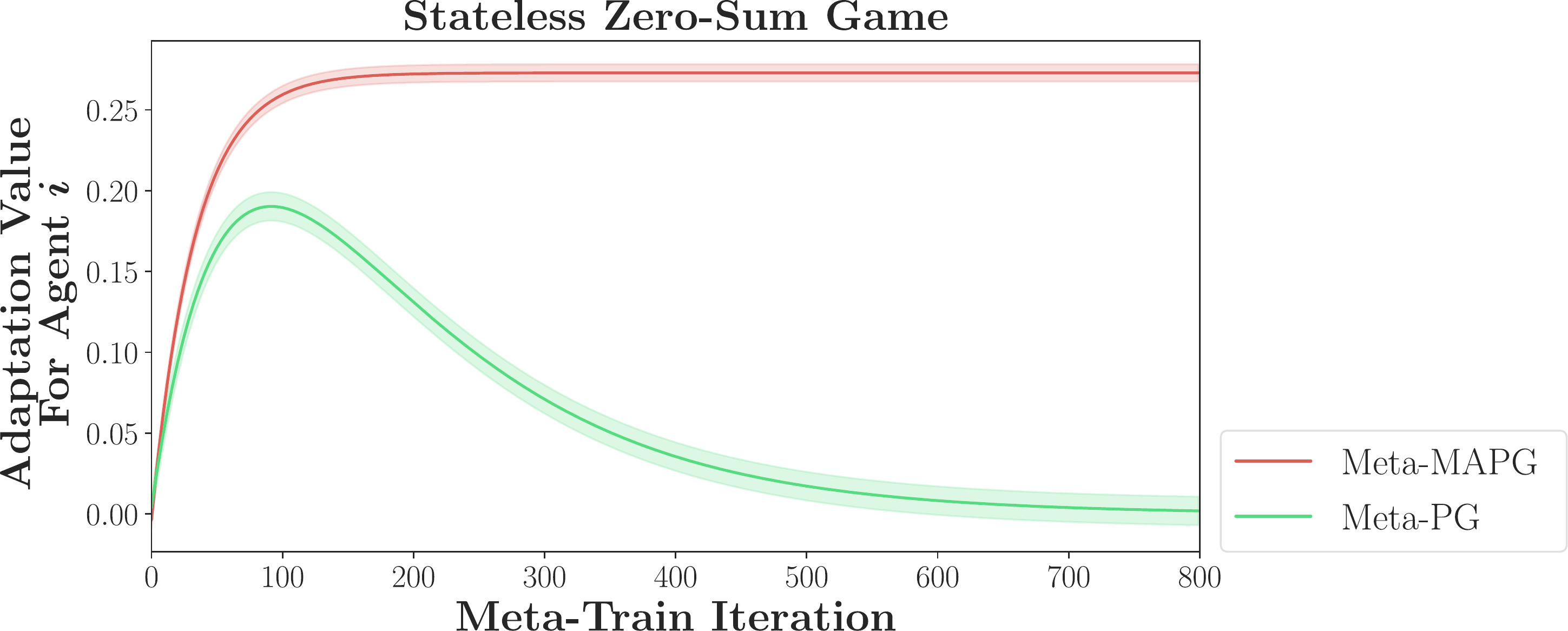}
    \vspace{-0.3cm}
    \caption{Meta-training result in the stateless zero-sum game when training a meta-agent $i$ with either Meta-PG~\citep{alshedivat2018continuous} or Meta-MAPG. Meta-MAPG achieves better adaptation than Meta-PG thanks to the peer learning gradient. Mean and 95\% confidence interval computed for 200 random samples are shown.
    }
    \label{fig:toy-example}
    \vskip-0.2in
\end{figure}

\textbf{Example 1.} \textit{Failure to consider our influence on the learning of other agents can result in biased and sometimes even counterproductive optimization.}

For example, consider a stateless zero-sum game played between two agents based off the one shown by \citet{letcher2018stable}.
A meta-agent $i$ and an opponent $j$ maximize simple value functions $V^{i}_{\bm{\phi}_{\bm{\ell}}}\!=\!\phi^{i}_{\ell}\phi^{j}_{\ell}$ and $V^{j}_{\bm{\phi}_{\bm{\ell}}}\!=\!-\phi^{i}_{\ell}\phi^{j}_{\ell}$ respectively, where $\phi^{i}_{\ell},\phi^{j}_{\ell}\!\in\!\mathbb{R}$. 
We assume a maximum Markov chain length $L$ of 1 and focus on meta-agent $i$'s adaptation performance $V^{i}_{\bm{\phi}_{\bm{1}}}$. 
Given an opponent's initial policy parameters randomly sampled from $p(\bm{\phi}^{\bm{-i}}_{\bm{0}})$, we compare training a meta-agent $i$ with either Meta-PG or Meta-MAPG.
As~\cref{fig:toy-example} shows, Meta-PG performs biased updates that actually make its performance worse over the course of meta-training. 
In contrast, by considering the opponent's learning process, Meta-MAPG corrects for this bias, displaying smooth convergence to a winning strategy.
See~\cref{sec:details-stateless-zero-sum} for more details, including the meta-gradient derivation.

\subsection{Algorithm}
We provide pseudo-code for Meta-MAPG in~\cref{alg:meta-train} for meta-training and~\cref{alg:meta-test} for meta-testing in~\Cref{sec:meta-mapg-algorithm}.
Note that Meta-MAPG is centralized during meta-training as it requires the policy parameters of other agents to compute the peer learning gradient. 
For settings where a meta-agent cannot access the policy parameters of other agents during meta-training, we provide a decentralized meta-training algorithm with opponent modeling, motivated by the approach used in~\citet{foerster17lola}, in~\cref{sec:opponent-modelling} that computes the peer learning gradient while leveraging only an approximation of the parameters of peer agents.
Once meta-trained in either case, the adaptation to new agents during meta-testing is purely decentralized such that the meta-agent can decide how to shape other agents with its own observations and rewards alone. 

%% file: related_work.tex
\section{Related Work}
The standard approach for addressing non-stationarity in MARL is to consider information about the other agents and reason about the effects of their joint actions~\citep{hernandezLealK17survey}.
The literature on opponent modeling, for instance, infers opponents' behaviors and conditions an agent's policy on the inferred behaviors of others~\citep{he16opponent-modeling,raileanu18opponent-modeling,grover18policy-representation}.
Studies regarding the centralized training with decentralized execution framework~\citep{lowe17maddpg,foerster2017counterfactual,yang18mean-field-marl,wen2018probabilistic}, which accounts for the behaviors of others through a centralized critic, can also be classified into this category.
While this body of work alleviates non-stationarity, it is generally assumed that each agent will have a stationary policy in the future.
Because other agents can have different behaviors in the future as a result of learning~\citep{foerster17lola}, this incorrect assumption can cause sample inefficient and improper adaptation.
In contrast, Meta-MAPG models each agent's learning process, allowing a meta-learning agent to adapt efficiently.

Our approach is also related to prior work that considers the learning of other agents in the environment. 
This includes~\citet{zhang10lookahead} who attempted to discover the best response adaptation to the anticipated future policy of other agents. Our work is also related, as discussed previously, to LOLA \citep{foerster17lola} and more recent improvements~\citep{foerster2018dice}.
Another relevant idea explored by~\citet{letcher2018stable} is to interpolate between the frameworks of~\citet{zhang10lookahead} and~\citet{foerster17lola} in a way that guarantees convergence while influencing the opponent's future policy.
However, all of these approaches only account for the learning processes of other agents and fail to consider an agent's own non-stationary policy dynamics as in the own learning gradient discussed in the previous section. 
Additionally, these papers do not leverage meta-learning. As a result, these approaches may require many samples to properly adapt to new agents.

Meta-learning \citep{schmidhuber87meta_learning, bengio1992optimization} has recently become very popular as a method for improving sample efficiency in the presence of changing tasks in the deep RL literature \citep{wang2016learning,duan2016rl,maml,snail,Reptile}. See~\citet{vilalta2002perspective} and~\citet{hospedales2020meta} for in-depth surveys of meta-learning. In particular, our work builds on the popular model-agnostic meta-learning framework~\citep{maml}, where gradient-based learning is used both for conducting so called inner-loop learning and to improve this learning by computing gradients through the computational graph. When we train our agents so that the inner loop can accommodate for a dynamic Markov chain of other agent policies, we are leveraging an approach that has recently become popular for supervised learning called meta-continual learning~\citep{MER,Javed2019Meta,spigler2019meta,beaulieu2020learning,caccia2020online,Gupta2020LaMAMLLM}. 
This means that our agent trains not just to adapt to a single set of policies during meta-training, but rather to adapt to a set of changing policies with Markovian updates. 
As a result, we avoid an issue of past work \citep{alshedivat2018continuous} that required the use of importance sampling during meta-testing (see~\Cref{sec:meta-pg-difference-details} for more details).

%% file: experiment.tex
\section{Experiments}\label{sec:experiments}
We demonstrate Meta-MAPG's efficacy on a diverse suite of domains, including the full spectrum of mixed incentive, competitive, and cooperative settings. 
The code is available at \url{https://git.io/JZsfg}, and video highlights are available at \url{http://bit.ly/2L8Wvch}. 
The mean and 95\% confidence interval computed for 10 seeds are shown in each figure.

\subsection{Experimental Setup for Meta-Learning in MARL}
\begin{figure}[t]
    \centering
    \includegraphics[height=3.7cm]{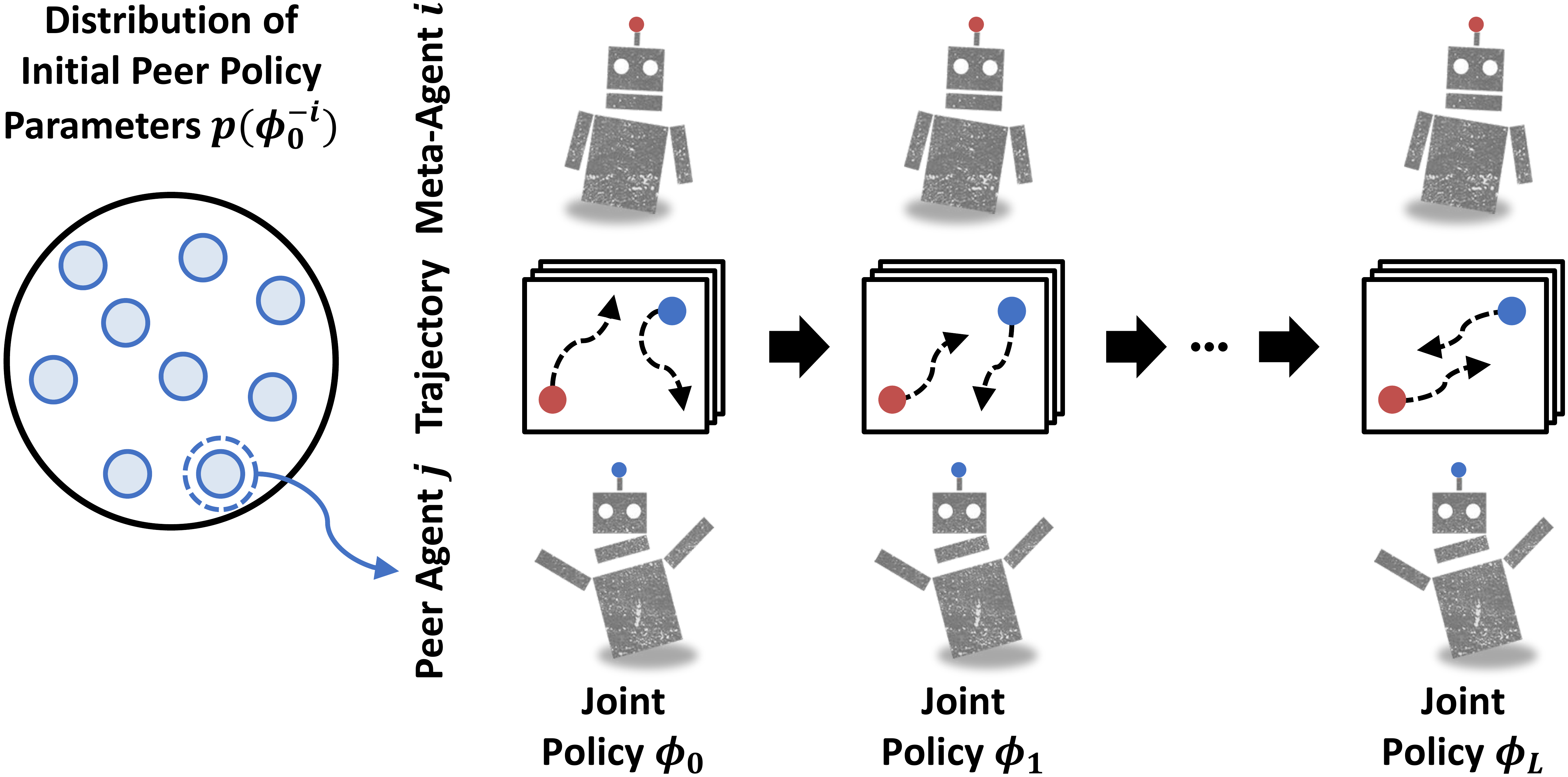}
    \vspace{-0.3cm}
    \caption{Experimental setup for meta-learning. A peer $j$'s policy is initialized randomly from a population $p(\bm{\phi}^{\bm{-i}}_{\bm{0}})$ and is updated based on the policy gradient optimization, requiring a meta-agent $i$ to adapt to differently initialized $j$ and its changing policy.}
    \label{fig:meta-learning-setup}
    \vskip-0.16in
\end{figure}

\noindent\header{Training protocol.}
In our experiments, we evaluate a meta-agent $i$'s adaptability with respect to a peer agent $j$ learning in the environment. 
Specifically, agent $j$ is sampled from a population of initial peer policy parameters $p(\bm{\phi}^{\bm{-i}}_{\bm{0}})$. Then, both agents adapt their behavior following the policy gradient with a linear feature baseline~\citep{duan-16-linearbaseline} (see~\cref{fig:meta-learning-setup}). 
Importantly, the population captures diverse behaviors of $j$, such as varying expertise in solving a task, and $j$'s initial policy is hidden to $i$. 
Hence, a meta-agent $i$ should: 1) adapt to a differently initialized agent $j$ with varying behaviors and 2) continuously adapt with respect to $j$'s changing policy.

During meta-training, a meta-agent $i$ interacts with peers drawn from the distribution and optimizes its initial policy parameters $\phi^{i}_{0}$.
At the end of meta-training, a new peer $j$ is sampled from the distribution, and we measure $i$'s performance throughout the Markov chain.

\noindent\header{Adaptation baselines.}
We compare Meta-MAPG with the following baseline adaptation strategies for an agent $i$:
\begin{enumerate}[wide, labelindent=0pt, label=\arabic*), topsep=0pt]
    \itemsep 0pt
    \item Meta-PG~\citep{alshedivat2018continuous}: A meta-learning approach that only considers how to improve its own learning. We detail our implementation of Meta-PG and a low-level difference with the original method in~\cref{sec:meta-pg-difference-details}.
    \item LOLA-DiCE~\citep{foerster2018dice}: 
    An approach that only considers how to shape the learning dynamics of other agents in the environment through the Differentiable Monte-Carlo Estimator (DiCE) operation. Note that LOLA-DiCE is an extension of the original LOLA approach.
    \item REINFORCE~\citep{Williams1992}: A simple policy gradient approach that considers neither an agent's own learning nor the learning processes of other agents. This baseline represents multiagent approaches that assume each agent leverages a stationary policy in the future. 
\end{enumerate}

\noindent\header{Implementation.}
We implement each adaptation method's policy leveraging an LSTM and use the generalized advantage estimation~\citep{schulmanetal-16-gae} with a learned value function for the meta-optimization. 
We also apply DiCE during the inner-loop optimization~\citep{foerster2018dice} to compute the peer learning gradients, and we learn dynamic inner-loop learning rates during meta-training as suggested in~\citet{alshedivat2018continuous}.
We refer to Appendices E, F, G, I for the remaining details including hyperparameters.

\begin{figure}[t!]
\captionsetup[subfigure]{skip=0pt}
    \begin{subfigure}[b]{0.49\linewidth}
        \centering
        \footnotesize
        \resizebox{0.9\textwidth}{!}{
        \setlength{\tabcolsep}{2pt}
        \begin{tabular}[b]{cc|cc}
        \multicolumn{2}{c}{} & \multicolumn{2}{c}{Agent $j$}\\
        \parbox[t]{3mm}{\multirow{3}{*}{\rotatebox[origin=c]{90}{Agent $i$}}} 
            &       & $C$       & $D$       \\\cline{2-4}
        \rule{0pt}{3pt}    & $C$   & $(0.5,0.5)$ & $(\text{-}1.5,1.5)$  \\
            & $D$   & $(1.5,\text{-}1.5)$  & $(\text{-}0.5,\text{-}0.5)$ 
        \end{tabular}}
        \caption{IPD payoff table}
        \label{tab:ipd-payoff-table}
    \end{subfigure}
    \begin{subfigure}[b]{0.49\linewidth}
        \centering
        \footnotesize
        \resizebox{0.9\textwidth}{!}{
        \setlength{\tabcolsep}{2pt}
        \begin{tabular}{cc|ccc}
        \multicolumn{2}{c}{} & \multicolumn{3}{c}{Agent $j$}\\
        \parbox[t]{3mm}{\multirow{3}{*}{\rotatebox[origin=c]{90}{Agent $i$}}} 
            &       & $R$       & $P$ & $S$      \\\cline{2-5}
        \rule{0pt}{3pt}    & $R$   & $(0,0)$   & $(\text{-}1,1)$ & $(1,\text{-}1)$ \\
            & $P$   & $(1,\text{-}1)$  & $(0,0)$ & $(\text{-}1,1)$ \\
            & $S$   & $(\text{-}1,1)$  & $(1,\text{-}1)$ & $(0,0)$
        \end{tabular}}
        \caption{RPS payoff table}
        \label{tab:rps-payoff-table}
    \end{subfigure}
    \begin{subfigure}[b]{\linewidth}
        \centering
        \includegraphics[trim=100 0 0 0,clip, height=1.35cm]{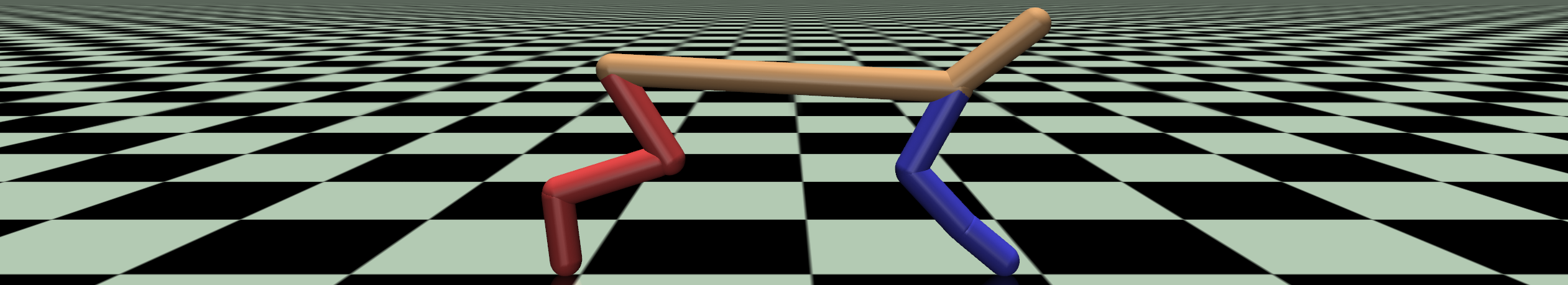}
        \caption{2-Agent HalfCheetah Benchmark}
        \label{fig:multi-mujoco-domain}
    \end{subfigure}
    \vspace{-0.74cm}
    \caption{\textbf{(a)} IPD payoff table. \textbf{(b)} RPS payoff table. \textbf{(c)} $2$-Agent HalfCheetah benchmark, where two agents are coupled within the robot and control the robot together: the red and blue agent control three joints of the back and front leg, respectively.}
    \vskip-0.2in
\end{figure}

\noindent\header{Experiment complexity.} 
We study various aspects of Meta-MAPG based on two repeated matrix games (see \Cref{tab:ipd-payoff-table,tab:rps-payoff-table}). 
We note that the repeated matrix domains we consider match the environment complexity of previous work on the topic of peer learning awareness in MARL~\citep{foerster17lola,foerster2018dice,letcher2018stable}.
We also considered the RoboSumo domain from~\citet{alshedivat2018continuous}. 
However, the training performance was very sensitive to the relative weights between the shaped reward functions and we could not reproduce the results from~\citet{alshedivat2018continuous} with the default weights.
Instead, we experimented with another challenging environment leveraging the multi-agent MuJoCo benchmark~\citep{dewitt2020deep_multimujoco} to demonstrate the scability of Meta-MAPG.
Specifically, the environment shown in~\Cref{fig:multi-mujoco-domain} has high complexity: 1) the domain has continuous and large observation/action spaces, and 2) the agents are coupled within the same robot, resulting in a more challenging control problem than controlling the robot alone with full autonomy.

\begin{figure*}[t!]
\captionsetup[subfigure]{skip=0pt, aboveskip=9pt}
    \begin{subfigure}[b]{0.247\linewidth}
        \centering
        \includegraphics[height=2.9cm]{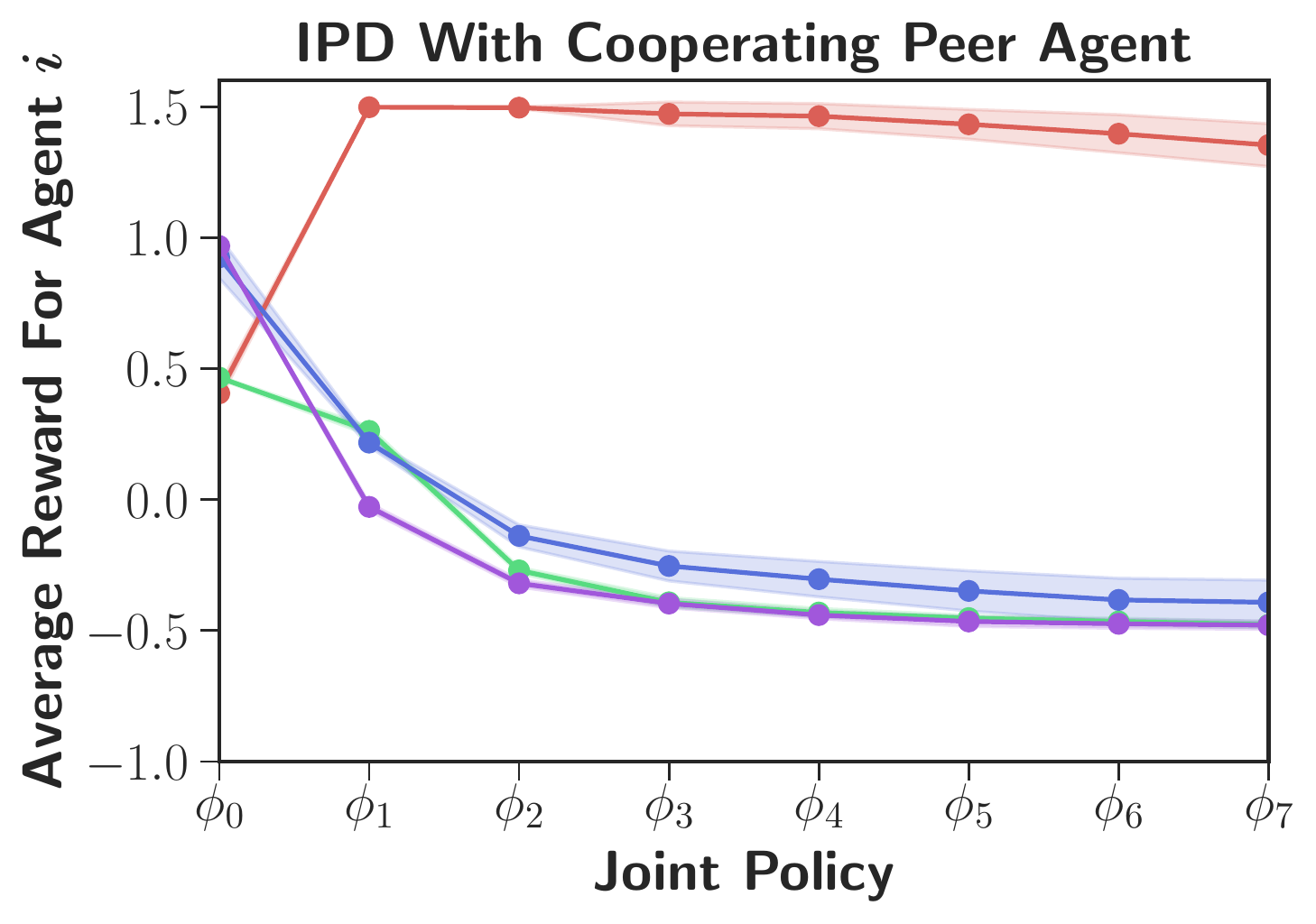}
        \caption{}
        \label{fig:ipd-cooperative-result}
    \end{subfigure}
    \begin{subfigure}[b]{0.247\linewidth}
        \centering
        \includegraphics[height=2.9cm]{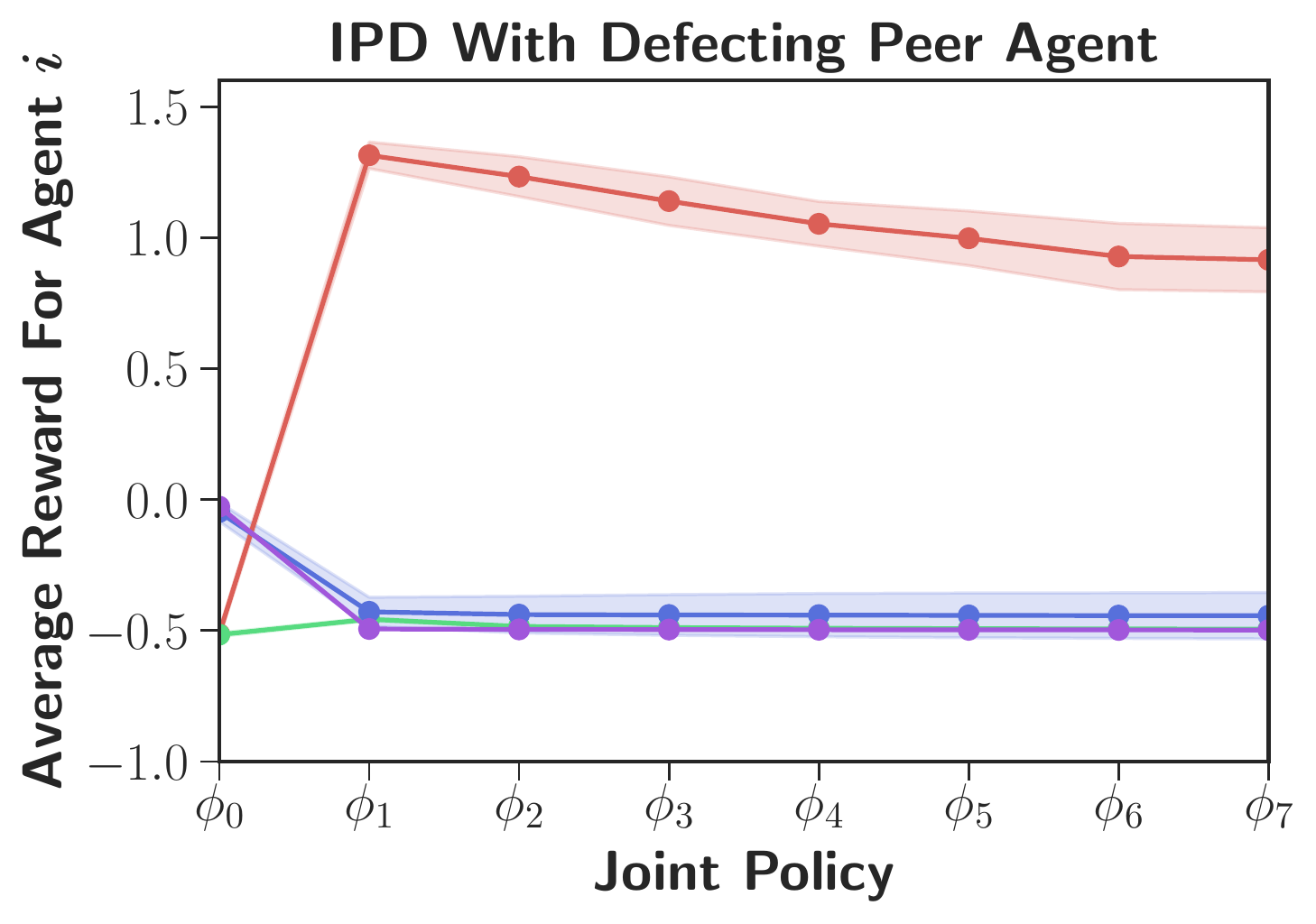}
        \caption{}
        \label{fig:ipd-defective-result}
    \end{subfigure}
    \begin{subfigure}[b]{0.247\linewidth}
        \centering
        \includegraphics[height=2.9cm]{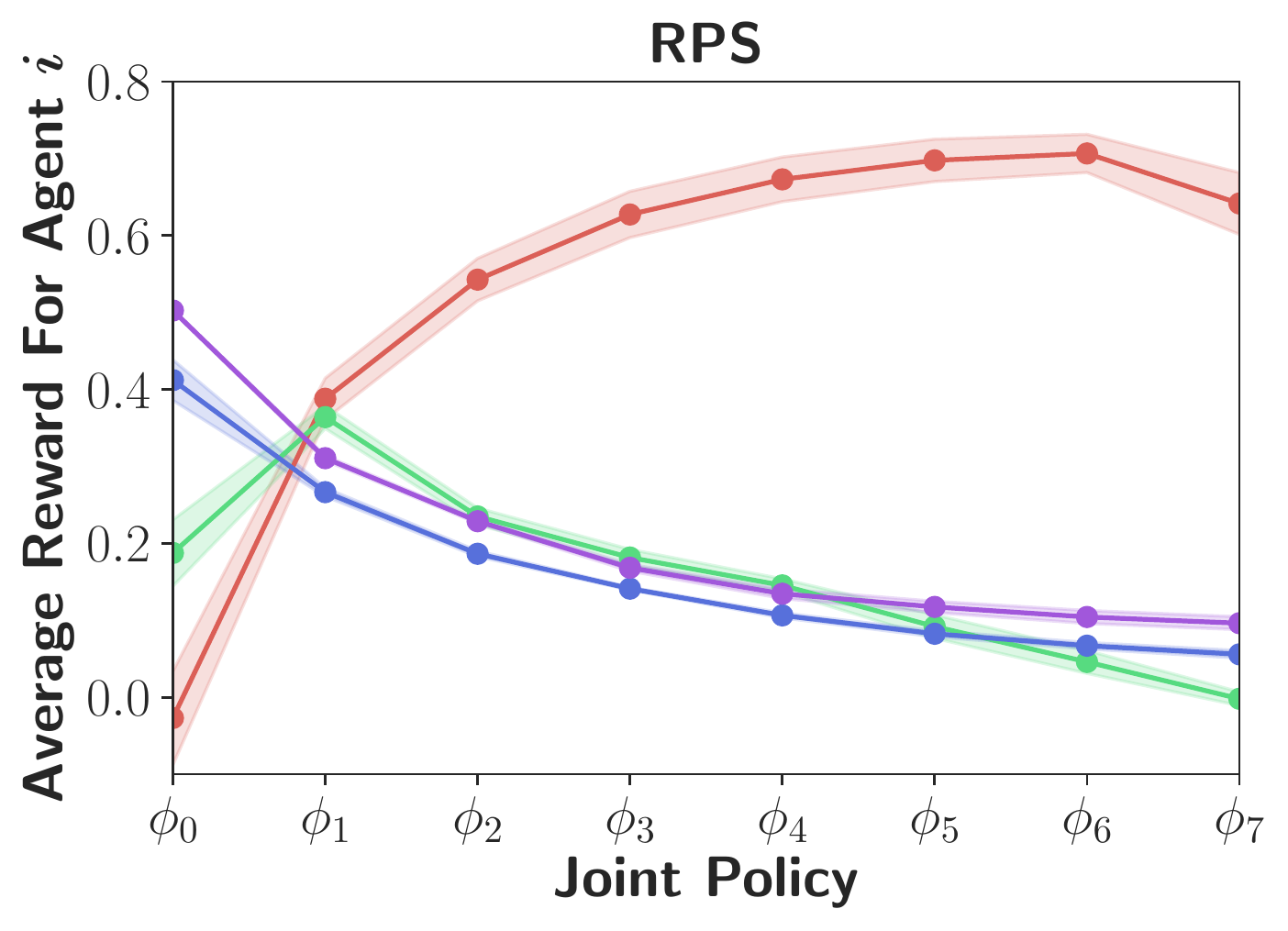}
        \caption{}
        \label{fig:rps-result}
    \end{subfigure}
    \begin{subfigure}[b]{0.247\linewidth}
        \centering
        \includegraphics[height=2.9cm]{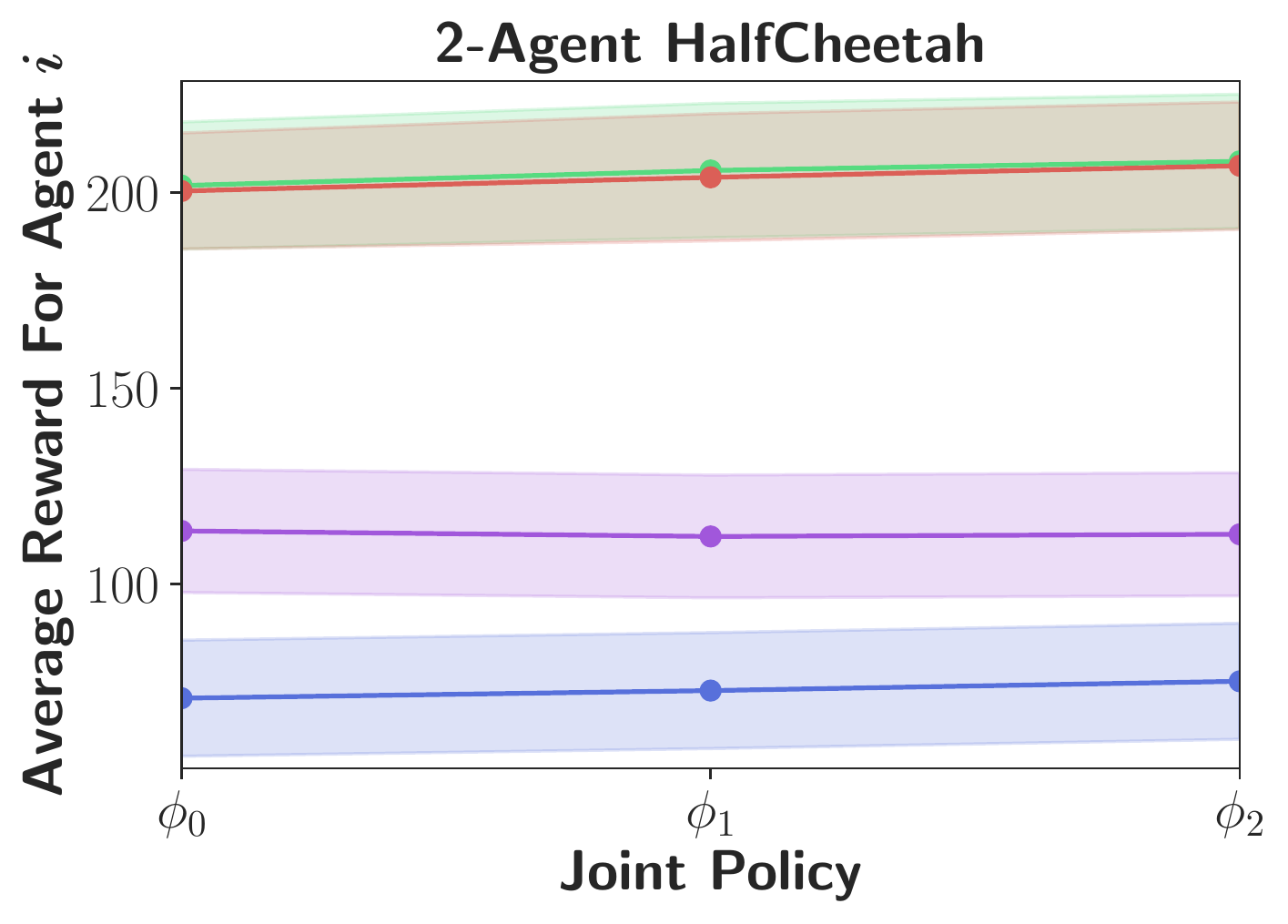}
        \caption{}
        \label{fig:halfcheeta-result-finale}
    \end{subfigure}
    
    \vspace{-0.9cm}
    \begin{subfigure}[b]{\linewidth}
        \centering
        \includegraphics[height=0.13cm]{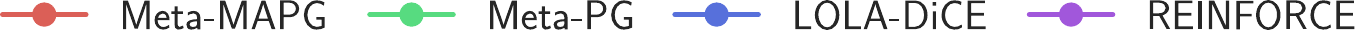}
    \end{subfigure}
    \caption{Adaptation performance during meta-testing in mixed incentive ((\textbf{a}) and (\textbf{b})), competitive (\textbf{c}), and cooperative (\textbf{d}) environments. 
    The results show that Meta-MAPG can successfully adapt to a new and learning peer agent throughout the Markov chain. 
    }
    \vskip-0.17in
\end{figure*}

\subsection{IPD Environment: Mixed Incentive}
\begin{question}
Is it essential to consider both an agent's own learning and the learning of others? \\[-2em]
\end{question}
To address this question, we consider the classic iterated prisoner's dilemma (IPD) domain. 
In IPD, agents $i$ and $j$ act by either (C)ooperating or (D)efecting and receive rewards according to the mixed incentive payoff defined in~\cref{tab:ipd-payoff-table}. 
As in~\citet{foerster17lola}, we model the state space as $s_{0}\!=\!\varnothing$ and $s_{t}\!=\!\bm{a}_{\bm{t-1}}$ for $t\geq1$.
For meta-learning, we construct a population of initial personas $p(\bm{\phi}^{\bm{-i}}_{\bm{0}})$ that include cooperating personas (i.e., having a probability of cooperating between $0.5$ and $1.0$ at any state) and defecting personas (i.e., having a probability of cooperating between $0$ and $0.5$ at any state).
\cref{fig:ipd-pca} in Appendix shows the population distribution utilized for training and evaluation.

The adaptation performance during meta-testing when an agent $i$, meta-trained with either Meta-MAPG or the baseline methods, interacts with an initially cooperating or defecting agent $j$ is shown in~\Cref{fig:ipd-cooperative-result} and~\Cref{fig:ipd-defective-result}, respectively.
In both cases, our meta-agent successfully infers the underlying persona of $j$ and adapts throughout the Markov chain obtaining higher rewards than the baselines. 
We observe that performance generally decreases as the number of joint policy updates increases across all adaptation methods. 
This decrease in performance is expected as each model is playing with another reasonable agent that is also constantly learning. 
As a result, $j$ realizes it could potentially achieve more reward by defecting more often.
Hence, to achieve good adaptation performance in IPD, an agent $i$ should attempt to shape $j$'s future policies toward staying cooperative as long as possible such that $i$ can take advantage, which is achieved by accounting for both an agent's own learning and the learning of other peer agents in Meta-MAPG. 

We explore each adaptation method in more detail by visualizing the action probability dynamics throughout the Markov chain. In general, we observe that the baseline methods have converged to initially defecting strategies, attempting to get larger rewards than a peer agent $j$ in the first trajectory $\tau_{\bm{\phi}_{\bm{0}}}$.
While this strategy can result in better initial performance than $j$, the peer agent will quickly change its policy so that it is defecting with high probability as well (see~\cref{fig:ipd-meta-pg-analysis,fig:ipd-dice-analysis,fig:ipd-reinforce-analysis} in Appendix). 
By contrast, our meta-agent learns to act cooperatively in $\tau_{\bm{\phi}_{\bm{0}}}$ and then take advantage by deceiving agent $j$ as it attempts to cooperate at future steps (see~\cref{fig:ipd-meta-mapg-analysis} in Appendix).

\begin{figure*}[t!]
\captionsetup[subfigure]{skip=0pt, aboveskip=0pt}
    \begin{subfigure}[b]{0.247\linewidth}
        \centering
        \includegraphics[height=2.9cm]{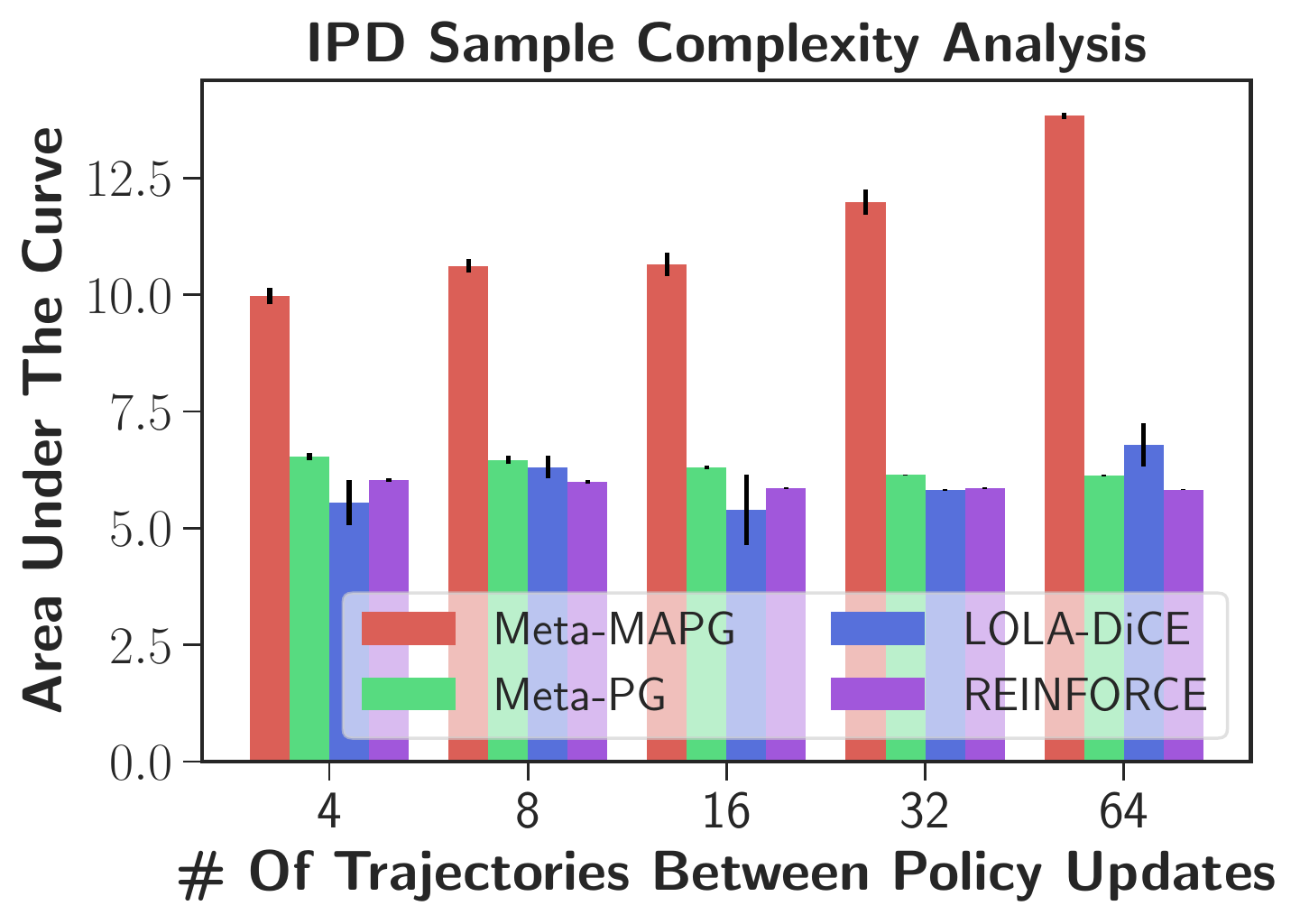}
        \caption{}
        \label{fig:sample-complexity}
    \end{subfigure}
    \begin{subfigure}[b]{0.247\linewidth}
        \centering
        \includegraphics[height=2.9cm]{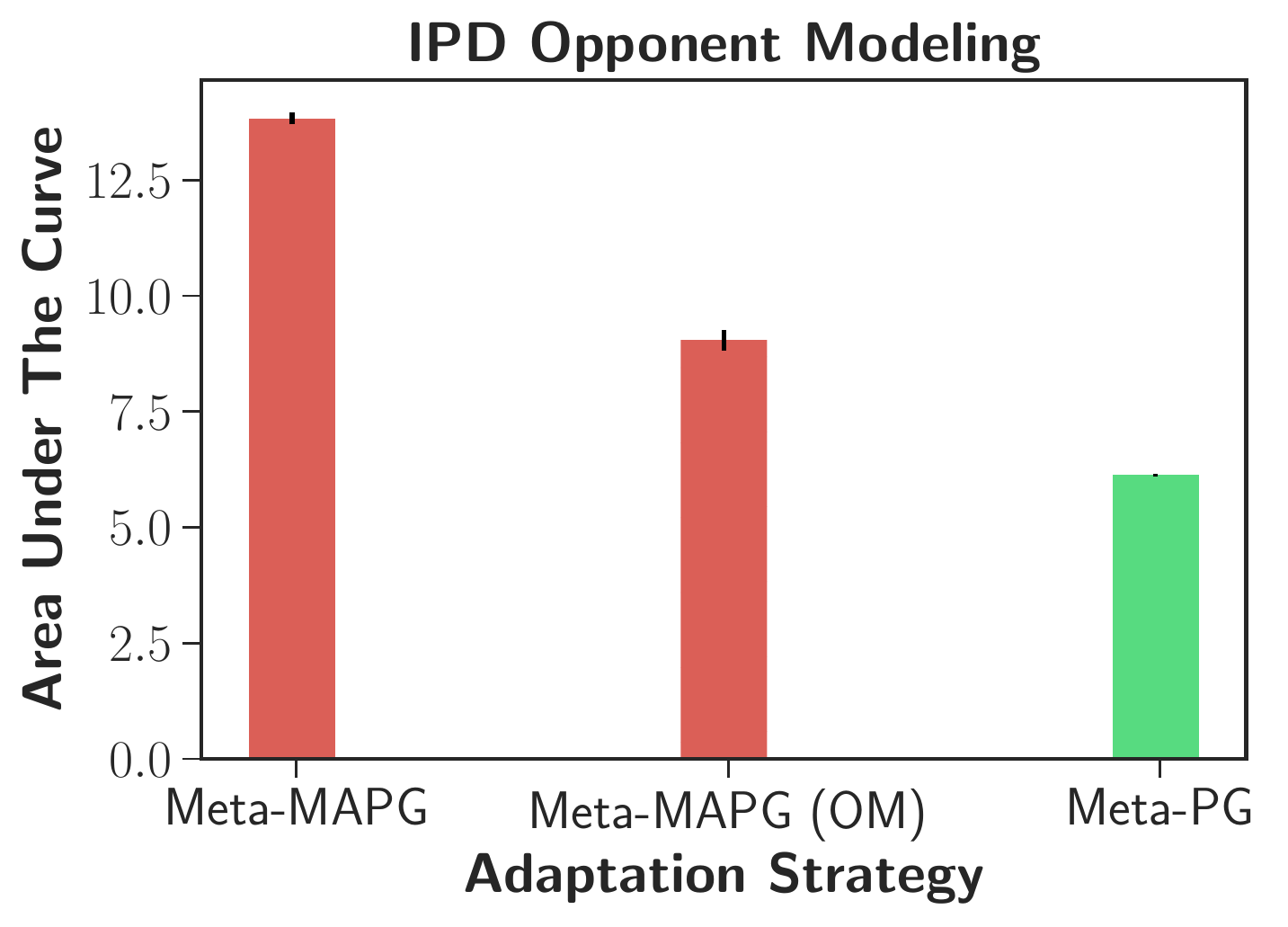}
        \caption{}
        \label{fig:ipd-om}
    \end{subfigure}
    \begin{subfigure}[b]{0.247\linewidth}
        \centering
        \includegraphics[height=2.9cm]{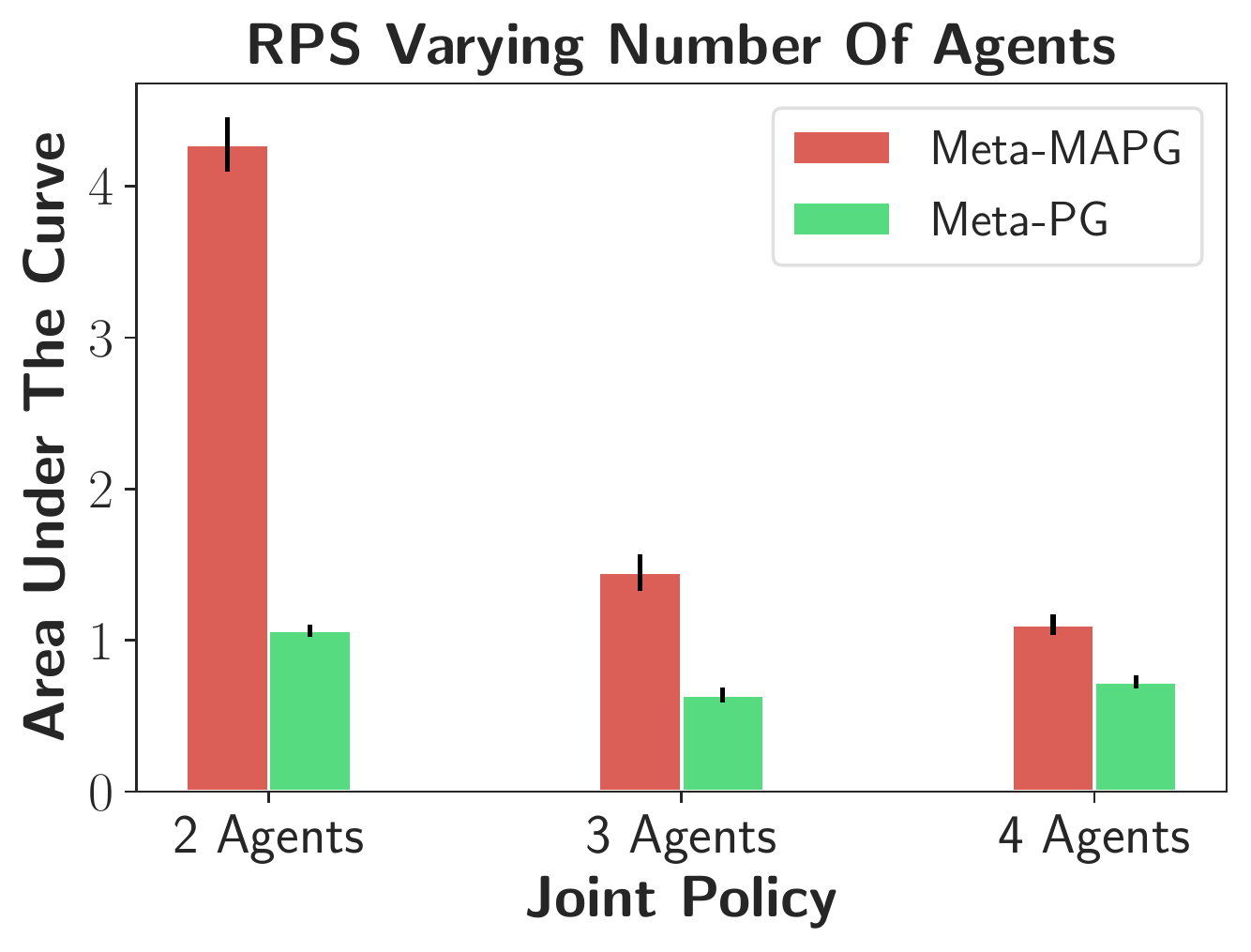}
        \caption{}
        \label{fig:rps-number}
    \end{subfigure}
    \begin{subfigure}[b]{0.247\linewidth}
        \centering
        \includegraphics[height=2.9cm]{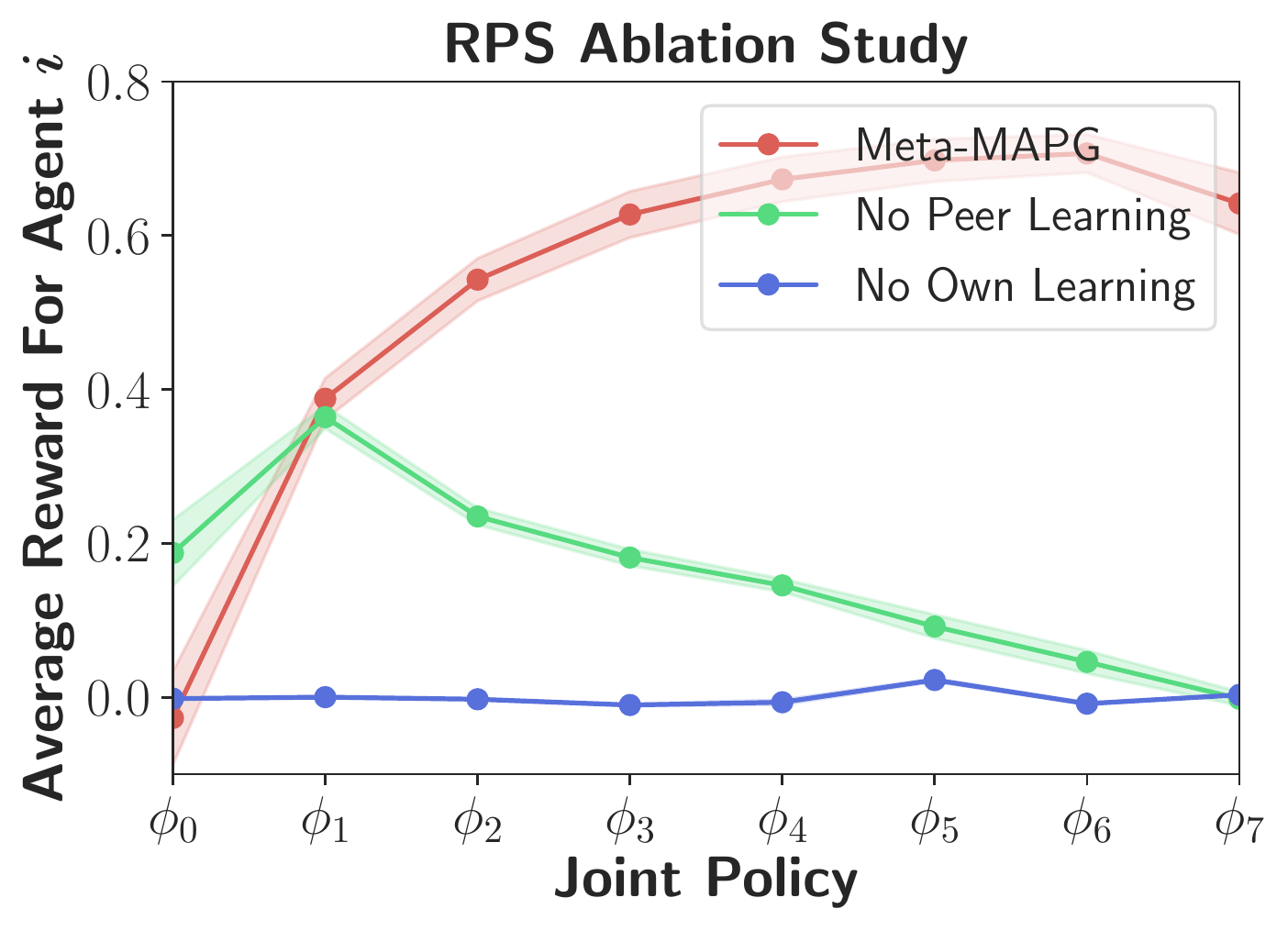}
        \caption{}
        \label{fig:rps-abb}
    \end{subfigure}
    \vspace{-0.9cm}
    \caption{
    (\textbf{a}) Adaptation performance with a varying number of trajectories. Meta-MAPG's performance generally improves with a larger $K$. (\textbf{b}) Adaptation performance with opponent modeling. Meta-MAPG with OM computes the peer learning gradient in a decentralized manner. (\textbf{c}) Adaptation performance with a varying number of agents, where Meta-MAPG achieves the best AUC in all cases.
    (\textbf{d}) Ablation study for Meta-MAPG. Meta-MAPG achieves significantly better performance than the two ablated baselines. 
    }
    \vskip-0.17in
\end{figure*}

\begin{question}
How is adaptation performance affected by the number of trajectories between changes? \\[-2em]
\end{question}
We control the level of non-stationarity by adjusting the number of trajectories $K$ between updates (refer to~\Cref{sec:markov-chain-of-policies}). 
The results in~\cref{fig:sample-complexity} shows that the area under the curve (AUC) (i.e., the reward summation during $\bm{\phi}_{\bm{1:L}}$) generally decreases when $K$ decreases in IPD. 
This result is expected since the inner-loop updates are based on the policy gradient, which can suffer from a high variance. 
Thus, with a smaller batch size, policy updates have a higher variance and lead to noisier policy updates. As a result, it is harder to anticipate and influence the future policies of peers. 
Nevertheless, Meta-MAPG achieves the best AUC in all cases.

\begin{question}
How does Meta-MAPG perform with decentralized meta-training?
\\[-2em]
\end{question}
We compare the performance of Meta-MAPG with and without opponent modeling (OM) in~\cref{fig:ipd-om}.
We note that Meta-MAPG with opponent modeling can infer policy parameters for peer agents and compute the peer learning gradient in a decentralized manner, performing better than the Meta-PG baseline.
However, opponent modeling introduces noise in predicting the future policy parameters of peer agents because the parameters must be inferred by observing the actions they take alone without any supervision about the parameters themselves. 
Thus, as expected, a meta-agent experiences difficulty in correctly considering the learning process of a peer, which leads to lower performance than Meta-MAPG with centralized meta-training.

\begin{question}
Can Meta-MAPG generalize its learning outside the meta-training distribution?  \\[-2em]
\end{question}
We have demonstrated that a meta-agent can generalize well and adapt to a new peer. 
However, we would like to investigate this further and see whether a meta-agent can still perform well when the meta-testing distribution is drawn from a significantly different distribution in IPD. 
We thus evaluate Meta-MAPG and Meta-PG using both in distribution (as in the previous questions) and out of distribution personas for $j$'s initial policies (see~\cref{fig:ipd-pca,fig:ipd-pca-ood} in Appendix).
Meta-MAPG achieves an AUC of $13.77\pm0.25$ and $11.12\pm0.33$ for the in and out of distribution evaluation, respectively. 
Meta-PG achieves an AUC of $6.13\pm0.05$ and $7.60\pm0.07$ for the in and out of distribution evaluation, respectively.
Variances are based on $5$ seeds and we leveraged $K\!=\!64$ for this experiment. 
We note that Meta-MAPG's performance decreases during the out of distribution evaluation, but still consistently performs better than the baseline.

\subsection{RPS Environment: Competitive}
\begin{question}
How effective is Meta-MAPG in a fully competitive scenario? \\[-2em]
\end{question}
We have demonstrated the benefit of our approach in the mixed incentive scenario of IPD. Here, we consider another classic iterated game, rock-paper-scissors (RPS) with a fully competitive payoff table (see \cref{tab:rps-payoff-table}). 
In RPS, at each time step agents $i$ and $j$ can choose an action of either (R)ock, (P)aper, or (S)cissors. The state space is defined as $s_{0}\!=\!\varnothing$ and $s_{t}\!=\!\bm{a}_{\bm{t-1}}$ for $t\geq1$.
For meta-learning, we consider a population of initial personas $p(\bm{\phi}^{\bm{-i}}_{\bm{0}})$, including the rock/paper/scissors persona (with a rock/paper/scissors actions probability between $1/3$ and $1$).

\cref{fig:rps-result} shows the adaptation performance during meta-testing.
Similar to the IPD results, we observe that the baseline methods have effectively converged to win against the opponent $j$ in the first few trajectories. For instance, agent $i$ has a high rock probability when playing against $j$ with a high initial scissors probability (see \cref{fig:rps-meta-pg-analysis,fig:rps-dice-analysis,fig:rps-reinforce-analysis} in Appendix). 
This strategy, however, results in the opponent quickly changing its behavior toward the mixed Nash equilibrium strategy of $(1/3, 1/3, 1/3)$ for the rock, paper, and scissors probabilities.
In contrast, our meta-agent learned to lose slightly in the first two trajectories $\tau_{\bm{\phi}_{\bm{0:1}}}$ to achieve much larger rewards in the later trajectories $\tau_{\bm{\phi}_{\bm{2:7}}}$ while relying on its ability to adapt more efficiently than its opponent (see~\cref{fig:rps-meta-mapg-analysis} in Appendix).
Compared to the IPD results, we observe that it is more difficult for our meta-agent to shape $j$'s future policies in RPS possibly due to the fact that RPS has a fully competitive payoff structure, while IPD has a mixed incentive structure.

\begin{question}
How effective is Meta-MAPG in settings with more than one peer?
\\[-2em]
\end{question}
Our meta-multiagent policy gradient theorem is general and can be applied to scenarios with more than one peer.
To validate this, we experiment with 3-player and 4-player RPS, where we consider sampling peers randomly from the entire persona population. 
We note that the domain becomes harder as the number of agents $n$ increases, because the state space dimension grows exponentially as a function of $n$ and a meta-agent needs to consider more peers in shaping their future policies. 
\Cref{fig:rps-number} shows a comparison against the Meta-PG baseline. 
We generally observe that the peer agents change their policies toward the mixed Nash equilibrium more quickly as the number of agents increases, which results in decreased performance for all methods. 
Nevertheless, Meta-MAPG achieves the best performance in all cases and can clearly be easily extended to settings with a greater number of agents.

\begin{question}
What happens if a meta-agent is trained without the own learning gradient?
\\[-2em]
\end{question}
Our meta-multiagent policy gradient theorem inherently includes both the own learning and peer learning gradient, but is it necessary to have both terms?
To answer this question, we conduct an ablation study and compare Meta-MAPG to two methods: one trained without the peer learning gradient and another trained without the own learning gradient. 
Note that not having the peer learning term is equivalent to Meta-PG, and not having the own learning term is similar to LOLA-DiCE but alternatively trained with a meta-optimization procedure. \Cref{fig:rps-abb} shows that a meta-agent trained without the peer learning term cannot properly exploit the peer agent's learning process. 
Also, a meta-agent trained without the own learning term cannot change its own policy effectively in response to anticipated learning by peer agents.
By contrast, Meta-MAPG achieves superior performance by accounting for both its own learning process and the learning processes of peer agents.

\subsection{2-Agent HalfCheetah Environment: Cooperative}
\begin{question}
Does considering the peer learning gradient always improve performance? \\[-2em]
\end{question}
To answer this question, we experiment with a fully cooperative setting of the the 2-Agent HalfCheetah domain~\citep{dewitt2020deep_multimujoco}, where the first and second agent control three joints of the back and front leg with continuous action spaces, respectively (see~\cref{fig:multi-mujoco-domain}). 
Both agents receive a joint reward corresponding to making the cheetah robot run to the right as soon as possible.
Note that two agents are coupled \textit{within} the cheetah robot, requiring close cooperation and coordination between them.

For meta-learning, we consider a population of teammates with varying degrees of expertise in running to the \textit{left} direction. 
Specifically, we pre-train a teammate $j$ and build a population based on checkpoints of its parameters during learning. 
\Cref{fig:mujoco-teammate-vis} in Appendix shows the distribution of $j$'s expertise. Note that we construct the distribution with a significant gap to ensure that the meta-testing distribution is sufficiently different from the meta-training/validation distribution. This allows us to effectively evaluate the ability of our approach to generalize its learning.
During meta-learning, $j$ is randomly initialized from this population of policies. 
Importantly, $j$ must adapt its behavior in this setting because the agent has achieved the \textit{opposite} skill during pre-training compared to the true objective of moving to the right. 
Hence, a meta-agent $i$ should succeed by both adapting to differently initialized teammates with varying expertise in moving the opposite direction, and guiding $j$'s learning process to coordinate the eventual right movement. 

Our results are displayed in~\cref{fig:halfcheeta-result-finale}. There are two notable observations. First, influencing peer learning does not help much in cooperative settings and Meta-MAPG performs similarly to Meta-PG.
The peer learning gradient attempts to shape the future policies of other agents so that the meta-agent can take advantage. In IPD, for example, the meta-agent influenced $j$ to be cooperative in the future such that the meta-agent can act with a high probability of the defect action and receive higher returns. 
However, in cooperative settings, due to the joint reward, $j$ is already changing its policies in order to benefit the meta-agent, resulting in a less significant effect with respect to the peer learning gradient. 
Second, Meta-PG and Meta-MAPG outperform the other approaches of LOLA-DiCE and REINFORCE, achieving higher rewards when interacting with a new teammate. 


%% file: conclusion.tex
\section{Conclusion}\label{sec:conclusion}
In this paper, we have introduced Meta-MAPG, a meta-learning algorithm that can adapt quickly to non-stationarity in the policies of other agents in a shared environment. The key idea underlying our proposed meta-optimization is to directly model both an agent's own learning process and the non-stationary policy dynamics of other agents.
We evaluated our method on several multiagent benchmarks and showed that Meta-MAPG is able to adapt more efficiently than previous state of the art approaches. 
We hope that our work can help provide the community with a theoretical foundation to build off for addressing the inherent non-stationarity of MARL in a principled manner. 

%% file: appendix.tex
\onecolumn
\section{Derivation of Meta-Multiagent Policy Gradient Theorem}\label{sec:proof-meta-multiagent-pg}
\textbf{Theorem 1.} (Meta-Multiagent Policy Gradient Theorem) \textit{For any stochastic game $\mathcal{M}_{n}$, the gradient of the meta-value function for agent $i$ at state $s_0$ with respect to current policy parameters $\phi_0^i$ evolving in the environment along with other peer agents using initial parameters $\bm{\phi}_{\bm{0}}^{\bm{-i}}$ is:}
\begin{align*}
\begin{split}
&\nabla_{\phi^{i}_{0}}V^{i}_{\bm{\phi}_{\bm{0:\ell+1}}}(s_0,\phi^{i}_{0})=\mathbb{E}_{p(\tau_{\bm{\phi}_{\bm{0:\ell}}}|\bm{\phi}_{\bm{0:\ell}})}\Big[\mathbb{E}_{p(\tau_{\bm{\phi}_{\bm{\ell+1}}}|\bm{\phi}_{\bm{\ell+1}})}\big[G^{i}(\tau_{\bm{\phi}_{\bm{\ell+1}}})\\
&\;\;\big(\underbrace{\nabla_{\phi^{i}_{0}}\log\pi(\tau_{\bm{\phi}_{\bm{0}}}|\phi^{i}_{0})}_{\text{Current Policy}}+\underbrace{\smallsum\nolimits_{\ell'=0}^{\ell}\nabla_{\phi^{i}_{0}}\log\pi(\tau_{\bm{\phi}_{\bm{\ell'+1}}}|\phi^{i}_{\ell'+1})}_{\text{Own Learning}}+\underbrace{\smallsum\nolimits_{\ell'=0}^{\ell}\nabla_{\phi^{i}_{0}}\log\pi(\tau_{\bm{\phi}_{\bm{\ell'+1}}}|\bm{\phi}^{\bm{-i}}_{\bm{\ell'+1}})}_{\text{Peer Learning}}\big)\big]\Big].
\end{split}
\end{align*}
\begin{proof}
We begin our derivation from the meta-value function defined in~\Cref{eqn:meta-value-function} and expand it with the state-action value and joint actions, assuming the conditional independence between agents' actions~\citep{wen2018probabilistic}: 
\begin{align}\label{eqn:theorem-step1}
\begin{split}
V^{i}_{\bm{\phi}_{\bm{0:\ell+1}}}(s_0,\phi^{i}_{0})=
&\mathbb{E}_{{p(\tau_{\bm{\phi}_{\bm{0:\ell}}}}|\bm{\phi}_{\bm{0:\ell}})}\Big[\mathbb{E}_{\tau_{\bm{\phi}_{\bm{\ell+1}}}}\big[G^{i}(\tau_{\bm{\phi}_{\bm{\ell+1}}})\big]\Big]=\mathbb{E}_{{p(\tau_{\bm{\phi}_{\bm{0:\ell}}}}|\bm{\phi}_{\bm{0:\ell}})}\Big[V^{i}_{\bm{\phi}_{\bm{\ell+1}}}(s_0)\Big]\\
=&\mathbb{E}_{{p(\tau_{\bm{\phi}_{\bm{0:\ell}}}}|\bm{\phi}_{\bm{0:\ell}})}\Big[\smallsum_{a^{i}_{0}}\pi(a^{i}_{0}|s_0,\phi^{i}_{\ell+1})\smallsum_{\bm{a}^{\bm{-i}}_{\bm{0}}}\bm{\pi}(\bm{a}^{\bm{-i}}_{\bm{0}}|s_{0},\bm{\phi}^{\bm{-i}}_{\bm{\ell+1}}) Q^{i}_{\bm{\phi}_{\bm{\ell+1}}}(s_{0},\bm{a}_{\bm{0}})\Big],
\end{split}
\end{align}
where $Q^{i}_{\bm{\phi}_{\bm{\ell+1}}}(s_{0},\bm{a}_{\bm{0}})$ is the state-action value under the joint policy with parameters $\bm{\phi}_{\bm{\ell+1}}$ at state $s_{0}$ with joint action $\bm{a}_{\bm{0}}$.

In~\cref{eqn:theorem-step1}, we note that both $\bm{\phi}^{\bm{i}}_{\bm{1:\ell}}$ and $\bm{\phi}^{\bm{-i}}_{\bm{1:\ell}}$ depend on $\phi^{i}_{0}$. 
Considering the joint update from $\bm{\phi}_{\bm{0}}$ to $\bm{\phi}_{\bm{1}}$, for simplicity, we can write the gradients in the inner-loop (see \Cref{eqn:inner-loop}) based on the multiagent stochastic policy gradient theorem~\citep{wei18multiagent-softQ}:
\begin{align}\label{eqn:theorem-step2}
\begin{split}
\nabla_{\phi^{-i}_{0}}\mathbb{E}_{p(\tau_{\bm{\phi}_{\bm{0}}}|\bm{\phi}_{\bm{0}})}\Big[G^{i}(\tau_{\bm{\phi}_{\bm{0}}})\Big]&=\smallsum_{s}\rho_{\bm{\phi}_{\bm{0}}}(s)\smallsum_{a^{i}}\nabla_{\phi^{i}_{0}}\pi(a^{i}|s,\phi^{i}_{0})\smallsum_{\bm{a}^{\bm{-i}}}\bm{\pi}(\bm{a}^{\bm{-i}}|s,\bm{\phi}^{\bm{-i}}_{\bm{0}})Q^{i}_{\bm{\phi}_{\bm{0}}}(s,\bm{a}),\\
\nabla_{\bm{\phi}^{\bm{-i}}_{\bm{0}}}\mathbb{E}_{p(\tau_{\bm{\phi}_{\bm{0}}}|\bm{\phi}_{\bm{0}})}\Big[\bm{G}^{\bm{-i}}(\tau_{\bm{\phi}_{\bm{0}}})\Big]&=\smallsum_{s}\rho_{\bm{\phi}_{\bm{0}}}(s)\smallsum_{\bm{a}^{\bm{-i}}}\nabla_{\bm{\phi}^{\bm{-i}}_{\bm{0}}}\bm{\pi}(\bm{a}^{\bm{-i}}|s,\bm{\phi}^{\bm{-i}}_{\bm{0}})\smallsum_{a^{i}}\pi(a^{i}|s,\phi^{i}_{0})\bm{Q}^{\bm{-i}}_{\bm{\phi}_{\bm{0}}}(s,\bm{a}),
\end{split}
\end{align}
where $\rho_{\bm{\phi}_{\bm{0}}}$ denotes the stationary distribution under the joint policy with parameters $\bm{\phi}_{\bm{0}}$.
Importantly, the inner-loop gradients for an agent $i$ and its peers are a function of $\phi^{i}_{0}$. 
Hence, the updated joint policy parameter $\bm{\phi}_{\bm{1}}$ depends on $\phi^{i}_{0}$. 
Following~\Cref{eqn:theorem-step2}, the successive inner-loop optimization until $\bm{\phi}_{\bm{\ell+1}}$ results in dependencies between $\phi^{i}_{0}$ and $\bm{\phi}^{\bm{i}}_{\bm{1:\ell+1}}$ and between $\phi^{i}_{0}$ and $\bm{\phi}^{\bm{-i}}_{\bm{1:\ell+1}}$ (see~\cref{fig:graph-meta-mapg}).

Having identified which terms are dependent on $\phi^{i}_{0}$, we continue from~\cref{eqn:theorem-step1} and derive the gradient of the meta-value function with respect to $\phi^{i}_{0}$ by applying the product rule:
\begin{align}\label{eqn:theorem-step3}
\nabla_{\phi^{i}_{0}}V^{i}_{\bm{\phi}_{\bm{0:\ell+1}}}(s_0,\phi^{i}_{0})&=\nabla_{\phi^{i}_{0}}\Big[\mathbb{E}_{{p(\tau_{\bm{\phi}_{\bm{0:\ell}}}}|\bm{\phi}_{\bm{0:\ell}})}\Big[\smallsum_{a^{i}_{0}}\pi(a^{i}_{0}|s_0,\phi^{i}_{\ell+1})\smallsum_{\bm{a}^{\bm{-i}}_{\bm{0}}}\bm{\pi}(\bm{a}^{\bm{-i}}_{\bm{0}}|s_{0},\bm{\phi}^{\bm{-i}}_{\bm{\ell+1}}) Q^{i}_{\bm{\phi}_{\bm{\ell+1}}}(s_{0},\bm{a}_{\bm{0}})\Big]\Big]\nonumber\\
&=\nabla_{\phi^{i}_{0}}\Big[\smallsum_{\tau_{\bm{\phi}_{\bm{0:\ell}}}} p(\tau_{\bm{\phi}_{\bm{0:\ell}}}|\bm{\phi}^{\bm{i}}_{\bm{0:\ell}},\bm{\phi}^{\bm{-i}}_{\bm{0:\ell}})\smallsum_{a^{i}_{0}}\pi(a^{i}_{0}|s_0,\phi^{i}_{\ell+1})\smallsum_{\bm{a}^{\bm{-i}}_{\bm{0}}}\bm{\pi}(\bm{a}^{\bm{-i}}_{\bm{0}}|s_{0},\bm{\phi}^{\bm{-i}}_{\bm{\ell+1}}) Q^{i}_{\bm{\phi}_{\bm{\ell+1}}}(s_{0},\bm{a}_{\bm{0}})\Big]\nonumber\\
&=\underbrace{\nabla_{\phi^{i}_{0}}\Big[\smallsum_{\tau_{\bm{\phi}_{\bm{0:\ell}}}} p(\tau_{\bm{\phi}_{\bm{0:\ell}}}|\bm{\phi}^{\bm{i}}_{\bm{0:\ell}},\bm{\phi}^{\bm{-i}}_{\bm{0:\ell}})\Big]\smallsum_{a^{i}_{0}}\pi(a^{i}_{0}|s_0,\phi^{i}_{\ell+1})\smallsum_{\bm{a}^{\bm{-i}}_{\bm{0}}}\bm{\pi}(\bm{a}^{\bm{-i}}_{\bm{0}}|s_{0},\bm{\phi}^{\bm{-i}}_{\bm{\ell+1}}) Q^{i}_{\bm{\phi}_{\bm{\ell+1}}}(s_{0},\bm{a}_{\bm{0}})}_{\text{Term A}}+\nonumber\\
&\;\;\;\;\:\underbrace{\smallsum_{\tau_{\bm{\phi}_{\bm{0:\ell}}}} p(\tau_{\bm{\phi}_{\bm{0:\ell}}}|\bm{\phi}^{\bm{i}}_{\bm{0:\ell}},\bm{\phi}^{\bm{-i}}_{\bm{0:\ell}})\Big[\smallsum_{a^{i}_{0}}\nabla_{\phi^{i}_{0}}\pi(a^{i}_{0}|s_0,\phi^{i}_{\ell+1})\Big]\smallsum_{\bm{a}^{\bm{-i}}_{\bm{0}}}\bm{\pi}(\bm{a}^{\bm{-i}}_{\bm{0}}|s_{0},\bm{\phi}^{\bm{-i}}_{\bm{\ell+1}}) Q^{i}_{\bm{\phi}_{\bm{\ell+1}}}(s_{0},\bm{a}_{\bm{0}})}_{\text{Term B}}+\\
&\;\;\;\;\:\underbrace{\smallsum_{\tau_{\bm{\phi}_{\bm{0:\ell}}}} p(\tau_{\bm{\phi}_{\bm{0:\ell}}}|\bm{\phi}^{\bm{i}}_{\bm{0:\ell}},\bm{\phi}^{\bm{-i}}_{\bm{0:\ell}})\smallsum_{a^{i}_{0}}\pi(a^{i}_{0}|s_0,\phi^{i}_{\ell+1})\Big[\smallsum_{\bm{a}^{\bm{-i}}_{\bm{0}}}\nabla_{\phi^{i}_{0}}\bm{\pi}(\bm{a}^{\bm{-i}}_{\bm{0}}|s_{0},\bm{\phi}^{\bm{-i}}_{\bm{\ell+1}})\Big] Q^{i}_{\bm{\phi}_{\bm{\ell+1}}}(s_{0},\bm{a}_{\bm{0}})}_{\text{Term C}}+\nonumber\\
&\;\;\;\;\:\underbrace{\smallsum_{\tau_{\bm{\phi}_{\bm{0:\ell}}}} p(\tau_{\bm{\phi}_{\bm{0:\ell}}}|\bm{\phi}^{\bm{i}}_{\bm{0:\ell}},\bm{\phi}^{\bm{-i}}_{\bm{0:\ell}})\smallsum_{a^{i}_{0}}\pi(a^{i}_{0}|s_0,\phi^{i}_{\ell+1})\smallsum_{\bm{a}^{\bm{-i}}_{\bm{0}}}\bm{\pi}(\bm{a}^{\bm{-i}}_{\bm{0}}|s_{0},\bm{\phi}^{\bm{-i}}_{\bm{\ell+1}})\Big[\nabla_{\phi^{i}_{0}}Q^{i}_{\bm{\phi}_{\bm{\ell+1}}}(s_{0},\bm{a}_{\bm{0}})\Big]}_{\text{Term D}}\nonumber.
\end{align}
We first focus on the derivative of the trajectories $\tau_{\bm{\phi}_{\bm{0:\ell}}}$ in Term A:
\begin{align}\label{eqn:theorem-step4}
\begin{split}
\nabla_{\phi^{i}_{0}}\Big[\smallsum_{\tau_{\bm{\phi}_{\bm{0:\ell}}}} p(\tau_{\bm{\phi}_{\bm{0:\ell}}}|\bm{\phi}^{\bm{i}}_{\bm{0:\ell}},\bm{\phi}^{\bm{-i}}_{\bm{0:\ell}})\Big]&=\nabla_{\phi^{i}_{0}}\Big[\smallsum_{\tau_{\bm{\phi}_{\bm{0}}}} p(\tau_{\bm{\phi}_{\bm{0}}}|\phi^{i}_{0},\bm{\phi}^{\bm{-i}}_{\bm{0}})\smallsum_{\tau_{\bm{\phi}_{\bm{1}}}} p(\tau_{\bm{\phi}_{\bm{1}}}|\phi^{i}_{1},\bm{\phi}^{\bm{-i}}_{\bm{1}})\times\sdots\times\smallsum_{\tau_{\bm{\phi}_{\bm{\ell}}}} p(\tau_{\bm{\phi}_{\bm{\ell}}}|\phi^{i}_{\ell},\bm{\phi}^{\bm{-i}}_{\bm{\ell}})\Big]\\
&=\Big[\smallsum_{\tau_{\bm{\phi}_{\bm{0}}}}\nabla_{\phi^{i}_{0}} p(\tau_{\bm{\phi}_{\bm{0}}}|\phi^{i}_{0},\bm{\phi}^{\bm{-i}}_{\bm{0}})\Big]\textstyle\prod_{\forall\ell'\in\{0,\sdots,\ell\}\setminus \{0\}}\smallsum_{\tau_{\bm{\phi}_{\bm{\ell'}}}} p(\tau_{\bm{\phi}_{\bm{\ell'}}}|\phi^{i}_{\ell'},\bm{\phi}^{\bm{-i}}_{\bm{\ell'}})+\\
&\;\;\;\;\:\Big[\smallsum_{\tau_{\bm{\phi}_{\bm{1}}}}\nabla_{\phi^{i}_{1}} p(\tau_{\bm{\phi}_{\bm{1}}}|\phi^{i}_{1},\bm{\phi}^{\bm{-i}}_{\bm{1}})\Big]\textstyle\prod_{\forall\ell'\in\{0,\sdots,\ell\}\setminus \{1\}}\smallsum_{\tau_{\bm{\phi}_{\bm{\ell'}}}} p(\tau_{\bm{\phi}_{\bm{\ell'}}}|\phi^{i}_{\ell'},\bm{\phi}^{\bm{-i}}_{\bm{\ell'}})+\sdots+\\
&\;\;\;\;\:\Big[\smallsum_{\tau_{\bm{\phi}_{\bm{\ell}}}}\nabla_{\phi^{i}_{\ell}} p(\tau_{\bm{\phi}_{\bm{\ell}}}|\phi^{i}_{\ell},\bm{\phi}^{\bm{-i}}_{\bm{\ell}})\Big]\textstyle\prod_{\forall\ell'\in\{0,\sdots,\ell\}\setminus \{\ell\}}\smallsum_{\tau_{\bm{\phi}_{\bm{\ell'}}}} p(\tau_{\bm{\phi}_{\bm{\ell'}}}|\phi^{i}_{\ell'},\bm{\phi}^{\bm{-i}}_{\bm{\ell'}}),
\end{split}
\end{align}
where the probability of collecting a trajectory under the joint policy with parameters $\bm{\phi}_{\bm{\ell}}$ is given by:
\begin{align}\label{eqn:theorem-step5}
\begin{split}
p(\tau_{\bm{\phi}_{\bm{\ell}}}|\phi^{i}_{\ell},\bm{\phi}^{\bm{-i}}_{\bm{\ell}})=p(s_0)\textstyle\prod\nolimits_{t=0}^{H}\pi(a^{i}_{t}|s_{t},\phi^{i}_{\ell})\bm{\pi}(\bm{a}^{\bm{-i}}_{\bm{t}}|s_{t},\bm{\phi}^{\bm{-i}}_{\bm{\ell}})\mathcal{P}(s_{t+1}|s_{t},\bm{a}_{\bm{t}}).
\end{split}
\end{align}

Using~\cref{eqn:theorem-step5} and the log-derivative trick,~\cref{eqn:theorem-step4} can be further expressed as:
\begin{align}\label{eqn:theorem-step6}
\begin{split}
&\Big[\mathbb{E}_{p(\tau_{\bm{\phi}_{\bm{0}}}|\bm{\phi}_{\bm{0}})}\nabla_{\phi^{i}_{0}}\log\pi(\tau_{\bm{\phi}_{\bm{0}}}|\phi^{i}_{0})\Big]\textstyle\prod\nolimits_{\forall\ell'\in\{0,\sdots,\ell\}\setminus \{0\}}\smallsum_{\tau_{\bm{\phi}_{\bm{\ell'}}}}p(\tau_{\bm{\phi}_{\bm{\ell'}}}|\phi^{i}_{\ell'},\bm{\phi}^{\bm{-i}}_{\bm{\ell'}})+\\
&\Big[\mathbb{E}_{p(\tau_{\bm{\phi}_{\bm{1}}}|\bm{\phi}_{\bm{1}})}\nabla_{\phi^{i}_{0}}\Big(\log\pi(\tau_{\bm{\phi}_{\bm{1}}}|\phi^{i}_{1})+\log\bm{\pi}(\tau_{\bm{\phi}_{\bm{1}}}|\bm{\phi}^{\bm{-i}}_{\bm{1}})\Big)\Big]\textstyle\prod\nolimits_{\forall\ell'\in\{0,\sdots,\ell\}\setminus \{1\}}\smallsum_{\tau_{\bm{\phi}_{\bm{\ell'}}}}p(\tau_{\bm{\phi}_{\bm{\ell'}}}|\phi^{i}_{\ell'},\bm{\phi}^{\bm{-i}}_{\bm{\ell'}})+\sdots+\\
&\Big[\mathbb{E}_{p(\tau_{\bm{\phi}_{\bm{\ell}}}|\bm{\phi}_{\bm{\ell}})}\nabla_{\phi^{i}_{0}}\Big(\log\pi(\tau_{\bm{\phi}_{\bm{\ell}}}|\phi^{i}_{\ell})+\log\bm{\pi}(\tau_{\bm{\phi}_{\bm{\ell}}}|\bm{\phi}^{\bm{-i}}_{\bm{\ell}})\Big)\Big]\textstyle\prod\nolimits_{\forall\ell'\in\{0,\sdots,\ell\}\setminus \{\ell\}}\smallsum_{\tau_{\bm{\phi}_{\bm{\ell'}}}}p(\tau_{\bm{\phi}_{\bm{\ell'}}}|\phi^{i}_{\ell'},\bm{\phi}^{\bm{-i}}_{\bm{\ell'}})
\end{split}
\end{align}
where the summations of the log-terms, such as $\nabla_{\phi^{i}_{0}}\!\Big(\!\log\pi(\tau_{\bm{\phi}_{\bm{\ell}}}|\phi^{i}_{\ell})\!+\!\log\bm{\pi}(\tau_{\bm{\phi}_{\bm{\ell}}}|\bm{\phi}^{\bm{-i}}_{\bm{\ell}})\!\Big)$ are \textit{inherently} included due to the sequential dependencies between $\phi^{i}_{0}$ and $\bm{\phi}_{\bm{1:\ell}}$. We use the result of~\cref{eqn:theorem-step6} and organize terms to arrive at the following expression for Term A in~\Cref{eqn:theorem-step3}:
\begin{align}\label{eqn:theorem-step7}
\begin{split}
\mathbb{E}_{p(\tau_{\bm{\phi}_{\bm{0:\ell}}}|\bm{\phi}_{\bm{0:\ell}})}\Big[&\Big(\nabla_{\phi^{i}_{0}}\log\pi(\tau_{\bm{\phi}_{\bm{0}}}|\phi^{i}_{0})+\smallsum_{\ell'=0}^{\ell-1}\nabla_{\phi^{i}_{0}}\log\pi(\tau_{\bm{\phi_{\ell'+1}}}|\phi^{i}_{\ell'+1})+\smallsum_{\ell'=0}^{\ell-1}\nabla_{\phi^{i}_{0}}\log\pi(\tau_{\bm{\phi_{\ell'+1}}}|\bm{\phi}^{\bm{-i}}_{\bm{\ell'+1}})\Big)\times\\
&\;\smallsum_{a^{i}_{0}}\pi(a^{i}_{0}|s_0,\phi^{i}_{\ell+1})\smallsum_{\bm{a}^{\bm{-i}}_{\bm{0}}}\bm{\pi}(\bm{a}^{\bm{-i}}_{\bm{0}}|s_{0},\bm{\phi}^{\bm{-i}}_{\bm{\ell+1}}) Q^{i}_{\bm{\phi}_{\bm{\ell+1}}}(s_{0},\bm{a}_{\bm{0}})
\Big].
\end{split}
\end{align}
Coming back to Term B-D in~\Cref{eqn:theorem-step3}, repeatedly unrolling the derivative of the Q-function $\nabla_{\phi^{i}_{0}}Q^{i}_{\bm{\phi}_{\bm{\ell+1}}}(s_{0},\bm{a}_{\bm{0}})$ by following~\citet{Sutton:1998} yields:
\begin{align}\label{eqn:theorem-step8}
\begin{split}
&
\mathbb{E}_{{p(\tau_{\bm{\phi}_{\bm{0:\ell}}}}|\bm{\phi}_{\bm{0:\ell}})}\Big[\smallsum_{s}\rho_{\bm{\phi}_{\bm{\ell+1}}}(s)\smallsum_{a^{i}}\nabla_{\phi^{i}_{0}}\pi(a^{i}|s,\phi^{i}_{\ell+1})\smallsum_{\bm{a}^{\bm{-i}}}\bm{\pi}(\bm{a}^{\bm{-i}}|s,\bm{\phi}^{\bm{-i}}_{\bm{\ell+1}})Q^{i}_{\bm{\phi}+1}(s,\bm{a})\Big]+\\
&
\mathbb{E}_{{p(\tau_{\bm{\phi}_{\bm{0:\ell}}}|\bm{\phi}_{\bm{0:\ell}})}}
\Big[\smallsum_{s}\rho_{\bm{\phi}_{\bm{\ell+1}}}(s)\smallsum_{\bm{a}^{\bm{-i}}}\nabla_{\phi^{i}_{0}}\bm{\pi}(\bm{a}^{\bm{-i}}|s,\bm{\phi}^{\bm{-i}}_{\bm{\ell+1}})\smallsum_{a^{i}}\pi(a^{i}|s,\phi^{i}_{\ell+1})Q^{i}_{\bm{\phi}_{\bm{\ell+1}}}(s,\bm{a})\Big],
\end{split}
\end{align}
which adds the consideration of future joint policy $\bm{\phi}_{\bm{\ell+1}}$ to~\cref{eqn:theorem-step7}.
Finally, we summarize~\Cref{eqn:theorem-step7,eqn:theorem-step8} together and express in expectations:
\begin{align*}
\begin{split}
&\nabla_{\phi^{i}_{0}}V^{i}_{\bm{\phi}_{\bm{0:\ell+1}}}(s_0,\phi^{i}_{0})=\mathbb{E}_{p(\tau_{\bm{\phi}_{\bm{0:\ell}}}|\bm{\phi}_{\bm{0:\ell}})}\Big[\mathbb{E}_{p(\tau_{\bm{\phi}_{\bm{\ell+1}}}|\bm{\phi}_{\bm{\ell+1}})}\big[G^{i}(\tau_{\bm{\phi}_{\bm{\ell+1}}})\\
&\;\;\big(\underbrace{\nabla_{\phi^{i}_{0}}\log\pi(\tau_{\bm{\phi}_{\bm{0}}}|\phi^{i}_{0})}_{\text{Current Policy}}+\underbrace{\smallsum\nolimits_{\ell'=0}^{\ell}\nabla_{\phi^{i}_{0}}\log\pi(\tau_{\bm{\phi}_{\bm{\ell'+1}}}|\phi^{i}_{\ell'+1})}_{\text{Own Learning}}+\underbrace{\smallsum\nolimits_{\ell'=0}^{\ell}\nabla_{\phi^{i}_{0}}\log\pi(\tau_{\bm{\phi}_{\bm{\ell'+1}}}|\bm{\phi}^{\bm{-i}}_{\bm{\ell'+1}})}_{\text{Peer Learning}}\big)\big]\Big].
\end{split}
\end{align*}
\end{proof}

\newpage
\section{Derivation of Meta-Policy Gradient Theorem}\label{sec:proof-meta-pg}
\textbf{Remark 1.} \textit{
Meta-PG can be considered as a special case of Meta-MAPG when assuming that other agents' learning in the environment is independent of the meta-agent's behavior.}
\begin{proof}
The framework by~\citet{alshedivat2018continuous} makes the implicit assumption that there exist no sequential dependencies between the future parameters of other agents $\bm{\phi}^{\bm{-i}}_{\bm{1:L}}$ and $\phi^{i}_{0}$. 
This assumption implies that the peers' policy updates in \cref{eqn:theorem-step2} are not a function of a meta-agent's policy.
As a result, the gradients of the peers' log-terms with respect to $\phi^{i}_{0}$ in~\cref{sec:proof-meta-multiagent-pg}, such as $\nabla_{\phi^{i}_{0}}\!\Big(\log\bm{\pi}(\tau_{\bm{\phi}_{\bm{\ell}}}|\bm{\phi}^{\bm{-i}}_{\bm{\ell}})\!\Big)$, become zero. 
Specifically, Term A in~\Cref{eqn:theorem-step3} simplifies to:
\begin{align}\label{eqn:simple-theorem-step7}
\begin{split}
\mathbb{E}_{p(\tau_{\bm{\phi}_{\bm{0:\ell}}}|\bm{\phi}_{\bm{0:\ell}})}\Big[&\Big(\nabla_{\phi^{i}_{0}}\log\pi(\tau_{\bm{\phi}_{\bm{0}}}|\phi^{i}_{0})+\smallsum_{\ell'=0}^{\ell-1}\nabla_{\phi^{i}_{0}}\log\pi(\tau_{\bm{\phi_{\ell'+1}}}|\phi^{i}_{\ell'+1})\Big)\times\\
&\;\smallsum_{a^{i}_{0}}\pi(a^{i}_{0}|s_0,\phi^{i}_{\ell+1})\smallsum_{\bm{a}^{\bm{-i}}_{\bm{0}}}\bm{\pi}(\bm{a}^{\bm{-i}}_{\bm{0}}|s_{0},\bm{\phi}^{\bm{-i}}_{\bm{\ell+1}}) Q^{i}_{\bm{\phi}_{\bm{\ell+1}}}(s_{0},\bm{a}_{\bm{0}})
\Big].
\end{split}
\end{align}

Similarly, Term B-D in~\Cref{eqn:theorem-step3} become: 
\begin{align}\label{eqn:simple-theorem-step8}
\begin{split}
&
\mathbb{E}_{{p(\tau_{\bm{\phi}_{\bm{0:\ell}}}}|\bm{\phi}_{\bm{0:\ell}})}\Big[\smallsum_{s}\rho_{\bm{\phi}_{\bm{\ell+1}}}(s)\smallsum_{a^{i}}\nabla_{\phi^{i}_{0}}\pi(a^{i}|s,\phi^{i}_{\ell+1})\smallsum_{\bm{a}^{\bm{-i}}}\bm{\pi}(\bm{a}^{\bm{-i}}|s,\bm{\phi}^{\bm{-i}}_{\bm{\ell+1}})Q^{i}_{\bm{\phi}+1}(s,\bm{a})\Big].
\end{split}
\end{align}
Finally, summarizing~\Cref{eqn:simple-theorem-step7,eqn:simple-theorem-step8} together, and expressing in expectations results in Meta-PG:
\begin{align*}
\begin{split}
&\nabla_{\phi^{i}_{0}}V^{i}_{\bm{\phi}_{\bm{0:\ell+1}}}(s_0,\phi^{i}_{0})=\mathbb{E}_{p(\tau_{\bm{\phi}_{\bm{0:\ell}}}|\bm{\phi}_{\bm{0:\ell}})}\Big[\mathbb{E}_{p(\tau_{\bm{\phi}_{\bm{\ell+1}}}|\bm{\phi}_{\bm{\ell+1}})}\big[G^{i}(\tau_{\bm{\phi}_{\bm{\ell+1}}})\\
&\;\;\big(\underbrace{\nabla_{\phi^{i}_{0}}\log\pi(\tau_{\bm{\phi}_{\bm{0}}}|\phi^{i}_{0})}_{\text{Current Policy}}+\underbrace{\smallsum\nolimits_{\ell'=0}^{\ell}\nabla_{\phi^{i}_{0}}\log\pi(\tau_{\bm{\phi}_{\bm{\ell'+1}}}|\phi^{i}_{\ell'+1})}_{\text{Own Learning}}\big)\big]\Big].
\end{split}
\end{align*}
\end{proof}

\section{Stateless Zero-Sum Game Details}\label{sec:details-stateless-zero-sum}
\noindent\paragraph{Derivation of Meta-MAPG.}
In the stateless zero-sum game, a meta-agent $i$ and an opponent $j$ maximize simple value functions $V^{i}_{\bm{\phi}_{\bm{\ell}}}\!=\!\phi^{i}_{\ell}\phi^{j}_{\ell}$ and $V^{j}_{\bm{\phi}_{\bm{\ell}}}\!=\!-\phi^{i}_{\ell}\phi^{j}_{\ell}$ respectively, where $\phi^{i}_{\ell},\phi^{j}_{\ell}\!\in\!\mathbb{R}$. 
We note that this domain has the episode horizon $H$ of 1 with a deterministic and continuous action space, where the action is equivalent to the policy parameter: $a^{i}_{\ell}\!=\!\phi^{i}_{\ell}$ and $a^{j}_{\ell}\!=\!\phi^{j}_{\ell}$. Because there are no stochastic factors in this domain, the meta-value function defined in~\cref{eqn:meta-value-function} can be simplified without the expectations:
\begin{gather}
V^{i}_{\bm{\phi}_{\bm{0:\ell+1}}}(\phi^{i}_{0})=V^{i}_{\bm{\phi}_{\bm{\ell+1}}}=\phi^{i}_{\ell+1}\phi^{j}_{\ell+1}.
\end{gather}
Assuming the maximum chain length $L$ of 1 for clarity, the inner-loop updates from $\bm{\phi}_{\bm{0}}$ to $\bm{\phi}_{\bm{1}}$ are:
\begin{align}
\begin{split}
\phi^{i}_{1}&=\phi^{i}_{0}+\alpha\nabla_{\phi^{i}_{0}}V^{i}_{\bm{\phi}_{\bm{0}}}=\phi^{i}_{0}+\alpha\nabla_{\phi^{i}_{0}}\big[\phi^{i}_{0}\phi^{j}_{0}\big]=\phi^{i}_{0}+\alpha\phi^{j}_{0},\\
\phi^{j}_{1}&=\phi^{j}_{0}+\alpha\nabla_{\phi^{j}_{0}}V^{j}_{\bm{\phi}_{\bm{0}}}=\phi^{j}_{0}+\alpha\nabla_{\phi^{j}_{0}}\big[-\phi^{i}_{0}\phi^{j}_{0}\big]=\phi^{j}_{0}-\alpha\phi^{i}_{0},
\end{split}
\end{align}
where $\alpha$ is the inner-loop learning rate.
Then, the meta-multiagent policy gradient can be directly computed without using the log-derivative trick:
\begin{gather}
\nabla_{\phi^{i}_{0}}V^{i}_{\bm{\phi}_{\bm{0:1}}}=\nabla_{\phi^{i}_{0}}V^{i}_{\bm{\phi}_{\bm{1}}}=\nabla_{\phi^{i}_{0}}\big[\phi^{i}_{1}\phi^{j}_{1}\big]=\big[\nabla_{\phi^{i}_{0}}\phi^{i}_{1}\big]\phi^{j}_{1}+\big[\nabla_{\phi^{i}_{0}}\phi^{j}_{1}\big]\phi^{i}_{1}=\phi^{j}_{1}-\alpha\phi^{i}_{1}.\label{eqn:stateless-product-rule}
\end{gather}
During meta-training, the initial policy parameter $\phi^{i}_{0}$ will be updated with the outer-loop learning rate $\beta$:
\begin{gather}
\phi^{i}_{0}:=\phi^{i}_{0}+\beta\nabla_{\phi^{i}_{0}}V^{i}_{\bm{\phi}_{\bm{0:1}}}=\phi^{i}_{0}+\beta\big[\phi^{j}_{1}-\alpha\phi^{i}_{1}\big].
\end{gather}

\noindent\paragraph{Derivation of Meta-PG.}
For Meta-PG~\citep{alshedivat2018continuous}, the framework assumes that there is no dependency between $\phi^{i}_{0}$ and $\phi^{j}_{1}$. 
Thus, the term $\big[\nabla_{\phi^{i}_{0}}\phi^{j}_{1}\big]$ in~\cref{eqn:stateless-product-rule} becomes zero, resulting in the meta-update of:
\begin{gather}
\phi^{i}_{0}:=\phi^{i}_{0}+\beta\nabla_{\phi^{i}_{0}}V^{i}_{\bm{\phi}_{\bm{0:1}}}=\phi^{i}_{0}+\beta\phi^{j}_{1}.
\end{gather}

\noindent\paragraph{Hyperparameter.}
We used the following hyperparameters: 1) randomly sampled initial opponent policy parameter from $-1$ to $1$ (i.e., $p(\bm{\phi}^{\bm{-i}}_{\bm{0}})=[-1, 1]$), 2) $\alpha\!=\!0.75$, and 3) $\beta\!=\!0.01$.

\section{Meta-MAPG Algorithm}\label{sec:meta-mapg-algorithm}
\vspace{-0.5cm}
\begin{figure}[H]
\begin{minipage}[t]{0.49\textwidth}
\begin{algorithm}[H]
	\caption{Meta-Learning at Training Time}\label{alg:meta-train}  
	\small
	\begin{algorithmic}[1]
	    \REQUIRE $p(\bm{\phi}^{\bm{-i}}_{\bm{0}})$: Distribution over peer agents' initial policy parameters; $\bm{\alpha},\beta$: Learning rates 
	    \STATE Randomly initialize $\phi^{i}_{0}$
		\WHILE{$\phi^{i}_{0}$ has not converged}
		    \STATE Sample a meta-train batch of $\bm{\phi}^{\bm{-i}}_{\bm{0}}\sim p(\bm{\phi}^{\bm{-i}}_{\bm{0}})$
		    \FOR{each $\bm{\phi}^{\bm{-i}}_{\bm{0}}$}
		        \FOR{$\ell=0,\sdots,L$}
		            \STATE Sample and store trajectory $\tau_{\bm{\phi}_{\bm{\ell}}}\sim p(\tau_{\bm{\phi}_{\bm{\ell}}}|\bm{\phi}_{\bm{\ell}})$
		            \STATE Compute $\bm{\phi}_{\bm{\ell+1}}=f(\bm{\phi}_{\bm{\ell}},\tau_{\bm{\phi}_{\bm{\ell}}},\bm{\alpha})$ from
		            inner-loop optimization (\cref{eqn:inner-loop})
		        \ENDFOR
		    \ENDFOR
		    \STATE Update $\phi^{i}_{0}\leftarrow\phi^{i}_{0}+\beta\sum_{\ell=0}^{L-1}\nabla_{\phi^{i}_{0}}V^{i}_{\bm{\phi}_{\bm{0:\ell+1}}}(s_0,\phi^{i}_{0})$ based on \cref{eqn:meta-multiagent-policy-gradient}
		\ENDWHILE
	\end{algorithmic}
\end{algorithm}
\end{minipage}
\hfill
\begin{minipage}[t]{0.49\textwidth}
\begin{algorithm}[H]
	\caption{Meta-Learning at Execution Time}\label{alg:meta-test}  
	\small
	\begin{algorithmic}[1]
	    \REQUIRE $p(\bm{\phi}^{\bm{-i}}_{\bm{0}})$: Distribution over peer agents' initial policy parameters; $\bm{\alpha}$: Learning rates; Optimized $\phi^{i*}_{0}$
	    \STATE Initialize $\phi^{i}_{0}\leftarrow\phi^{i*}_{0}$
		\STATE Sample a meta-test batch of $\bm{\phi}^{\bm{-i}}_{\bm{0}}\sim p(\bm{\phi}^{\bm{-i}}_{\bm{0}})$
		\FOR{each $\bm{\phi}^{\bm{-i}}_{\bm{0}}$}
		   \FOR{$\ell=0,\sdots,L$}
		      \STATE Sample trajectory $\tau_{\bm{\phi}_{\bm{\ell}}}\sim p(\tau_{\bm{\phi}_{\bm{\ell}}}|\bm{\phi}_{\bm{\ell}})$
		      \STATE Compute $\bm{\phi}_{\bm{\ell+1}}=f(\bm{\phi}_{\bm{\ell}},\tau_{\bm{\phi}_{\bm{\ell}}},\alpha)$ from inner-loop optimization (\cref{eqn:inner-loop})
		  \ENDFOR
		\ENDFOR
	\end{algorithmic}
\end{algorithm}
\end{minipage}
\end{figure}
\vspace{-0.5cm}

\subsection{Meta-MAPG with Opponent Modeling}\label{sec:opponent-modelling}
\vspace{-0.5cm}
\begin{figure}[H]
\begin{minipage}[t]{0.49\textwidth}
\begin{algorithm}[H]
	\caption{Meta-Learning at Training Time with OM}\label{alg:meta-train-om}  
	\small
	\begin{algorithmic}[1]
	    \REQUIRE $p(\bm{\phi}^{\bm{-i}}_{\bm{0}})$: Distribution over peer agents' initial policy parameters; $\bm{\alpha},\bm{\hat{\alpha}}^{\bm{-i}},\bm{\hat{\eta}}^{\bm{-i}},\beta$: Learning rates 
	    \STATE Randomly initialize $\phi^{i}_{0}$
		\WHILE{$\phi^{i}_{0}$ has not converged}
		    \STATE Sample a meta-train batch of $\bm{\phi}^{\bm{-i}}_{\bm{0}}\sim p(\bm{\phi}^{\bm{-i}}_{\bm{0}})$
		    \FOR{each $\bm{\phi}^{\bm{-i}}_{\bm{0}}$}
		         \STATE Randomly initialize $\bm{\hat{\phi}}^{\bm{-i}}_{\bm{0}}$
		        \FOR{$\ell=0,\sdots,L$}
		            \STATE Sample and store trajectory $\tau_{\bm{\phi}_{\bm{\ell}}}\sim p(\tau_{\bm{\phi}_{\bm{\ell}}}|\bm{\phi}_{\bm{\ell}})$
		            \STATE Approximate $\bm{\hat{\phi}}^{\bm{-i}}_{\bm{\ell}}=f(\bm{\hat{\phi}}^{\bm{-i}}_{\bm{\ell}},\tau_{\bm{\phi}_{\bm{\ell}}},\bm{\hat{\eta}}^{\bm{-i}})$ using opponent modeling (\cref{alg:opponent-modeling})
		            \STATE Compute $\bm{\phi}_{\bm{\ell+1}}=f(\bm{\phi}_{\bm{\ell}},\tau_{\bm{\phi}_{\bm{\ell}}},\bm{\alpha})$ from inner-loop optimization (\cref{eqn:inner-loop})
		            \STATE Compute $\bm{\hat{\phi}}^{\bm{-i}}_{\bm{\ell+1}}=f(\bm{\hat{\phi}}^{\bm{-i}}_{\bm{\ell}},\tau_{\bm{\phi}_{\bm{\ell}}},\bm{\hat{\alpha}}^{\bm{-i}})$ from inner-loop optimization (\cref{eqn:inner-loop})
		        \ENDFOR
		    \ENDFOR
		    \STATE Update $\phi^{i}_{0}\leftarrow\phi^{i}_{0}+\beta\sum_{\ell=0}^{L-1}\nabla_{\phi^{i}_{0}}V^{i}_{\bm{\phi}_{\bm{0:\ell+1}}}(s_0,\phi^{i}_{0})$ based on \cref{eqn:meta-multiagent-policy-gradient} and $\bm{\hat{\phi}}^{\bm{-i}}_{\bm{1:L}}$
		\ENDWHILE
	\end{algorithmic}
\end{algorithm}
\end{minipage}
\hfill
\begin{minipage}[t]{0.49\textwidth}
\begin{algorithm}[H]
	\caption{Opponent Modeling}\label{alg:opponent-modeling}  
	\small
	\begin{algorithmic}[1]
	    \FUNCTION{Opponent Modeling$(\bm{\hat{\phi}}^{\bm{-i}}_{\bm{\ell}},\tau_{\bm{\phi}_{\bm{\ell}}},\bm{\hat{\eta}}^{\bm{-i}})$}
	    	\WHILE{$\bm{\hat{\phi}}^{\bm{-i}}_{\bm{\ell}}$ has not converged}
	    	    \STATE Compute log-likelihood $\mathcal{L}_{\text{likelihood}}=f(\bm{\hat{\phi}}^{\bm{-i}}_{\bm{\ell}},\tau_{\bm{\phi}_{\bm{\ell}}})$ based on~\cref{eqn:log-likelihood}
	    	    \STATE Update $\bm{\hat{\phi}}^{\bm{-i}}_{\bm{\ell}}\leftarrow\bm{\hat{\phi}}^{\bm{-i}}_{\bm{\ell}}+\bm{\hat{\eta}}^{\bm{-i}}\nabla_{\bm{\hat{\phi}}^{\bm{-i}}_{\bm{\ell}}}\mathcal{L}_{\text{likelihood}}$
	    	\ENDWHILE
	    	\STATE \textbf{return} $\bm{\hat{\phi}}^{\bm{-i}}_{\bm{\ell}}$
       \ENDFUNCTION
	\end{algorithmic}
\end{algorithm}
\end{minipage}
\end{figure}
\vspace{-0.5cm}
We explain Meta-MAPG with opponent modeling (OM) for settings where a meta-agent cannot access the policy parameters of its peers during meta-training.
Our decentralized meta-training method in~\cref{alg:meta-train-om} replaces the other agents' true policy parameters $\bm{\phi}^{\bm{-i}}_{\bm{1:L}}$ with inferred parameters $\bm{\hat{\phi}}^{\bm{-i}}_{\bm{1:L}}$ in computing the peer learning gradient.
Specifically, we follow~\citet{foerster17lola} for opponent modeling and estimate $\bm{\hat{\phi}}^{\bm{-i}}_{\bm{\ell}}$ from $\tau_{\bm{\phi}_{\bm{\ell}}}$ using log-likelihood $\mathcal{L}_{\text{likelihood}}$ (Line 8 in~\cref{alg:meta-train-om}): 
\begin{align}\label{eqn:log-likelihood}
\begin{split}
\mathcal{L}_{\text{likelihood}}=\sum_{t=0}^{H}\log\bm{\pi}^{\bm{-i}}(\bm{a}^{\bm{-i}}_{\bm{t}}|s_{t},\bm{\hat{\phi}}^{\bm{-i}}_{\bm{\ell}}),
\end{split}
\end{align}
where $s_{t},\bm{a}_{\bm{t}}^{\bm{-i}}\in\tau_{\bm{\phi}_{\bm{\ell}}}$.
A meta-agent can obtain $\bm{\hat{\phi}}^{\bm{-i}}_{\bm{1:L}}$ by iteratively applying the opponent modeling procedure until the maximum chain length of $L$.
We also apply the inner-loop update with the Differentiable Monte-Carlo Estimator (DiCE)~\citep{foerster2018dice} to the inferred policy parameters of peer agents (Line 10 in~\cref{alg:meta-train-om}). 
By applying DiCE, we can save the sequential dependencies between $\phi^{i}_{0}$ and updates to the policy parameters of peer agents $\bm{\hat{\phi}}^{\bm{-i}}_{\bm{1:L}}$ in a computation graph and compute the peer learning gradient efficiently via automatic-differentiation (Line 13 in~\cref{alg:meta-train-om}).

\section{Additional Implementation Details}\label{sec:implementation-details}

\subsection{Network Structure}
Our neural networks for the policy and value function consist of a fully-connected input layer with $64$ units followed by a single-layer LSTM with $64$ units and a fully-connected output layer. 
We reset the LSTM states to zeros at the beginning of trajectories and retain them until the end of episodes. 
The LSTM policy outputs a probability for the categorical distribution in the iterated games (i.e., IPD, RPS). 
For the 2-Agent HalfCheetah domain, the policy outputs a mean and variance for the Gaussian distribution. 
We empirically observe that no parameter sharing between the policy and value network results in more stable learning than sharing the network parameters.

\subsection{Optimization}\label{sec:optimization-details}
We detail additional important notes about our implementation:
\begin{itemize}[leftmargin=*, wide, labelindent=0pt, topsep=0pt]
    \itemsep0pt
    \item We apply the linear feature baseline~\citep{duan-16-linearbaseline} and generalized advantage estimation (GAE)~\citep{schulmanetal-16-gae} during the inner-loop and outer-loop optimization, respectively, to reduce the variance in the policy gradient. 
    
    \item We use DiCE~\citep{foerster2018dice} to compute the peer learning gradient efficiently. Specifically, we apply DiCE during the inner-loop optimization and save the sequential dependencies between $\phi^{i}_{0}$ and $\bm{\phi}^{\bm{-i}}_{\bm{1:L}}$ in a computation graph. 
    Because the computation graph has the sequential dependencies, we can compute the peer learning gradient by the backpropagation of the meta-value function via the automatic-differentiation toolbox.
    
    \item Learning from diverse peers can potentially cause conflicting gradients and unstable learning. 
    In IPD, for instance, a strategy to adapt against cooperating peers can be completely opposite to the adaptation strategy against defecting peers, resulting in conflicting gradients.
    To address this potential issue, we use the projecting conflicting gradients (PCGrad)~\citep{yu20pcgrad} during the outer-loop optimization. We also have tested the baseline methods with PCGrad.
    
    \item We use a distributed training to speed up the meta-optimization. 
    Each thread interacts with a Markov chain of policies until the chain horizon and then computes the meta-optimization gradients using~\cref{eqn:meta-multiagent-policy-gradient}. Then, similar to~\citet{mniha16A3C}, each thread asynchronously updates the shared meta-agent's policy and value network parameters. 
\end{itemize}

\section{Additional Baseline Details}\label{sec:baseline-details}
We train all adaptation methods based on a meta-training set until convergence. 
We then measure the adaptation performance on a meta-testing set using the best-learned policy determined by a meta-validation set.

\subsection{Meta-PG}\label{sec:meta-pg-difference-details}
We have improved the Meta-PG baseline itself beyond its implementation in the original work~\citep{alshedivat2018continuous} to further isolate the importance of the peer learning gradient term.
Specifically, compared to~\citet{alshedivat2018continuous}, we make the following theoretical contributions to build on:
\begin{enumerate}[leftmargin=*, wide, labelindent=0pt, topsep=0pt, label=\arabic*)]
    \itemsep0em 
    \item \textbf{Underlying problem statement:} \citet{alshedivat2018continuous} bases their problem formulation off that of multi-task / continual single-agent RL. In contrast, ours is based on a general stochastic game between $n$ agents~\citep{shapley53stochastic}.
    \item \textbf{A Markov chain of joint policies:} \citet{alshedivat2018continuous} treats an evolving peer agent as an external factor, resulting in the absence of the sequential dependencies between a meta-agent's current policy and the peer agents' future policies in the Markov chain. However, our important insight is that the sequential dependencies exist in general multiagent settings as the peer agents are also learning agents based on trajectories by interacting with a meta-agent (see~\cref{fig:graph-meta-mapg}). 
    \item \textbf{Meta-objective:} The meta-objective defined in~\citet{alshedivat2018continuous} is based on single-agent settings. In contrast, our meta-objective is based on general multiagent settings (see~\Cref{eqn:meta-value-function,eqn:inner-loop}).
    \item \textbf{Meta-optimization gradient:} Compared to Al-Shedivat et al. (2018), our meta-optimization gradient inherently includes the additional term of the peer learning gradient that considers how an agent can directly influence peer's learning.
    \item \textbf{Importance sampling:} Compared to~\citet{alshedivat2018continuous}, we avoid using the importance sampling during meta-testing by modifying the meta-value function. Specifically, the framework uses a meta-value function on a pair consecutive joint policies, denoted $V^{i}_{\bm{\phi}_{\bm{\ell:\ell+1}}}(s_0,\phi^{i}_{0})$, which assumes initializing every $\phi^{i}_{\ell}$ from $\phi^{i}_{0}$. However, as noted in~\citet{alshedivat2018continuous}, this assumption requires interacting with the same peers multiple times and is often impossible during meta-testing. To address this issue, the framework uses the importance sampling correction during meta-testing. However, the correction generally suffers from high variance~\citep{wang16acer}. As such, we effectively avoid using the correction by initializing from $\phi^{i}_{0}$ only once at the beginning of Markov chains for both meta-training and meta-testing. 
\end{enumerate}

\subsection{LOLA-DiCE}
We used an open-source PyTorch implementation for LOLA-DiCE: \url{https://github.com/alexis-jacq/LOLA_DiCE}.
We make minor changes to the code, such as adding the LSTM policy and value function.

\section{Additional Experiment Details}\label{sec:experiment-details}
\subsection{IPD}
\begin{figure}[H]
\captionsetup[subfigure]{skip=0pt, aboveskip=0pt}
    \begin{subfigure}[b]{0.45\linewidth}
        \centering
        \includegraphics[height=2.9cm]{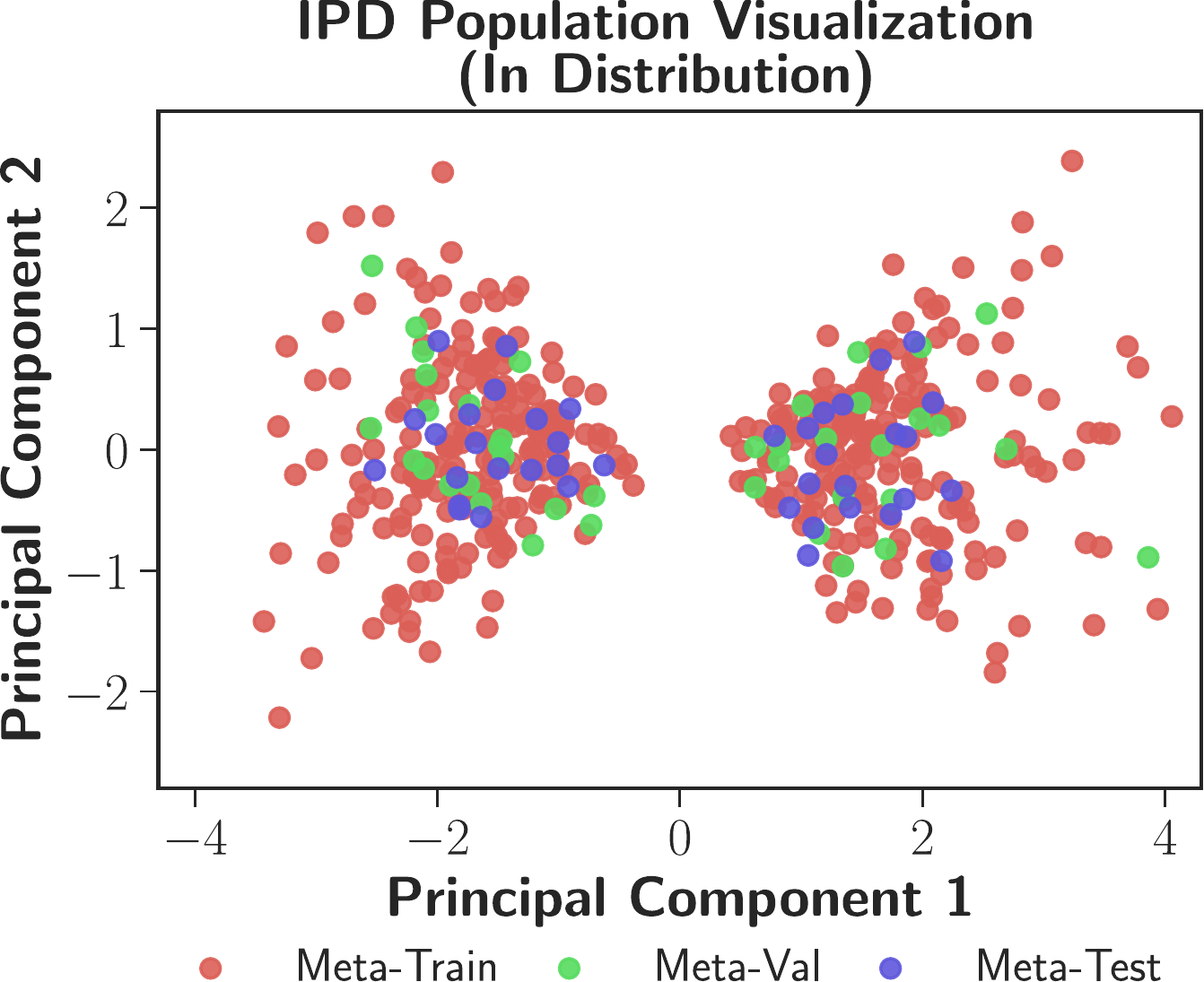}
        \caption{}
        \label{fig:ipd-pca}
    \end{subfigure}
    \begin{subfigure}[b]{0.45\linewidth}
        \centering
        \includegraphics[height=2.9cm]{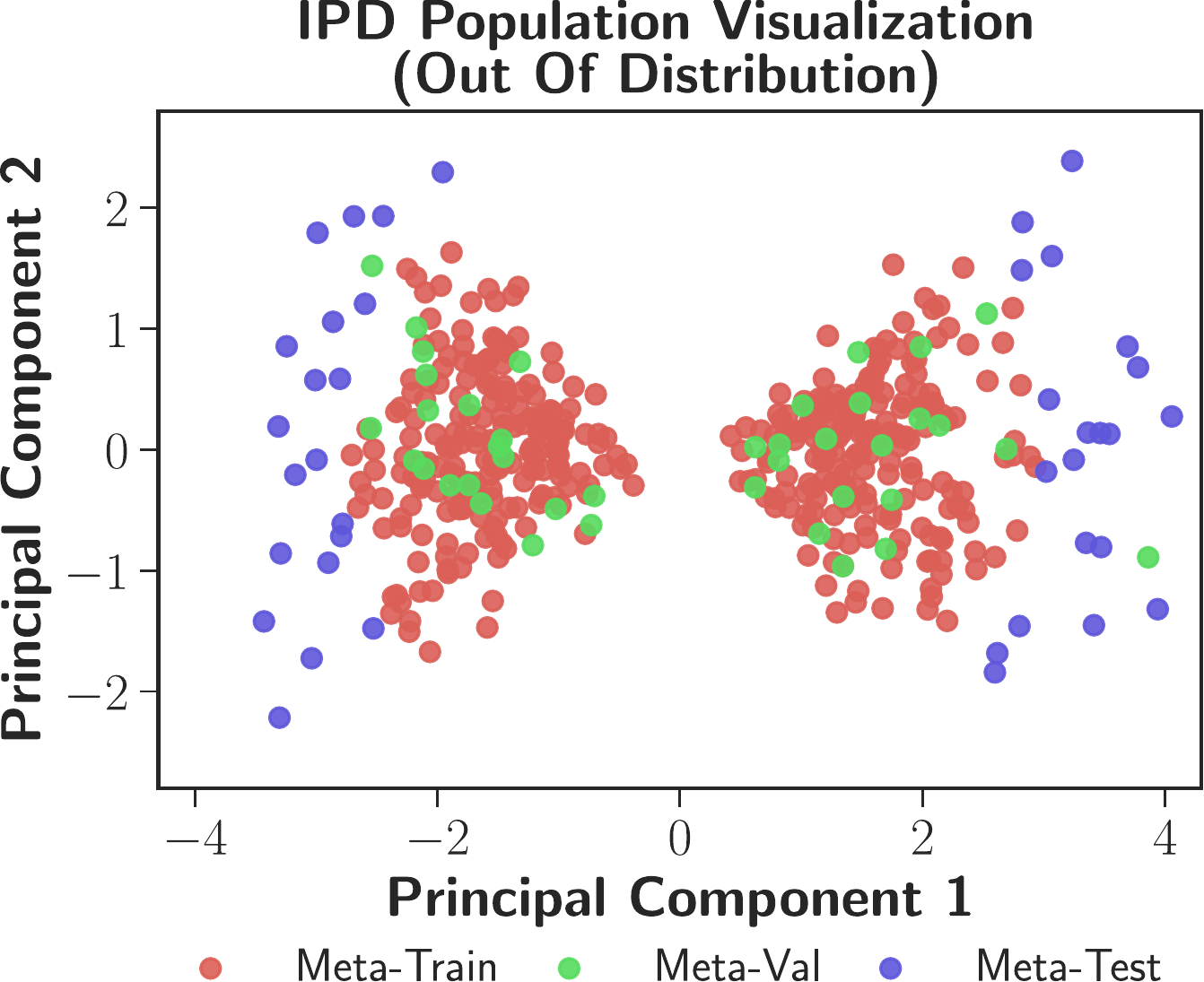}
        \caption{}
        \label{fig:ipd-pca-ood}
    \end{subfigure}
    \vspace{-0.4cm}
    \caption{
    (\textbf{a}) and (\textbf{b}) Visualization of $j$'s initial policy for in distribution and out of distribution meta-testing, respectively, where the out of distribution split has a smaller overlap between the policies used for meta-training/validation and those used for meta-testing.}
\end{figure}

We choose to represent the peer agent $j$'s policy as a tabular representation to effectively construct the population of initial personas $p(\bm{\phi}^{\bm{-i}}_{\bm{0}})$ for the meta-learning setup. 
Specifically, the tabular policy has a dimension of $5$ that corresponds to the number of states in IPD. 
Then, we randomly sample a probability between $0.5$ and $1.0$ and a probability between $0$ and $0.5$ at each state to construct the cooperating and defecting population, respectively. 
As such, the tabular representation enables us to sample as many as personas but also controllable distribution $p(\bm{\phi}^{\bm{-i}}_{\bm{0}})$ by merely adjusting the probability range.
We sample a total of $480$ initial personas, including cooperating personas and defecting personas, and split them into $400$ for meta-training, $40$ for meta-validation, and $40$ for meta-testing.
\Cref{fig:ipd-pca} and \Cref{fig:ipd-pca-ood} visualize in and out of distribution, respectively, where we used the principal component analysis (PCA) with two components. 
\vspace{-0.8cm}


\subsection{RPS}
In RPS, we follow the same meta-learning setup as in IPD, except we sample a total of $720$ initial opponent personas, including rock, paper, and scissors personas, and split them into $600$ for meta-training, $60$ for meta-validation, and $60$ for meta-testing. 
Additionally, because RPS has three possible actions, we sample a rock preference probability between $1/3$ and $1$ for building the rock persona population, where the rock probability is larger than the other two action probabilities. We follow the same procedure for constructing the paper and scissors persona population.


\subsection{2-Agent HalfCheetah}
\begin{wrapfigure}{r}{0.3\textwidth}
  \centering
  \includegraphics[width=0.75\linewidth]{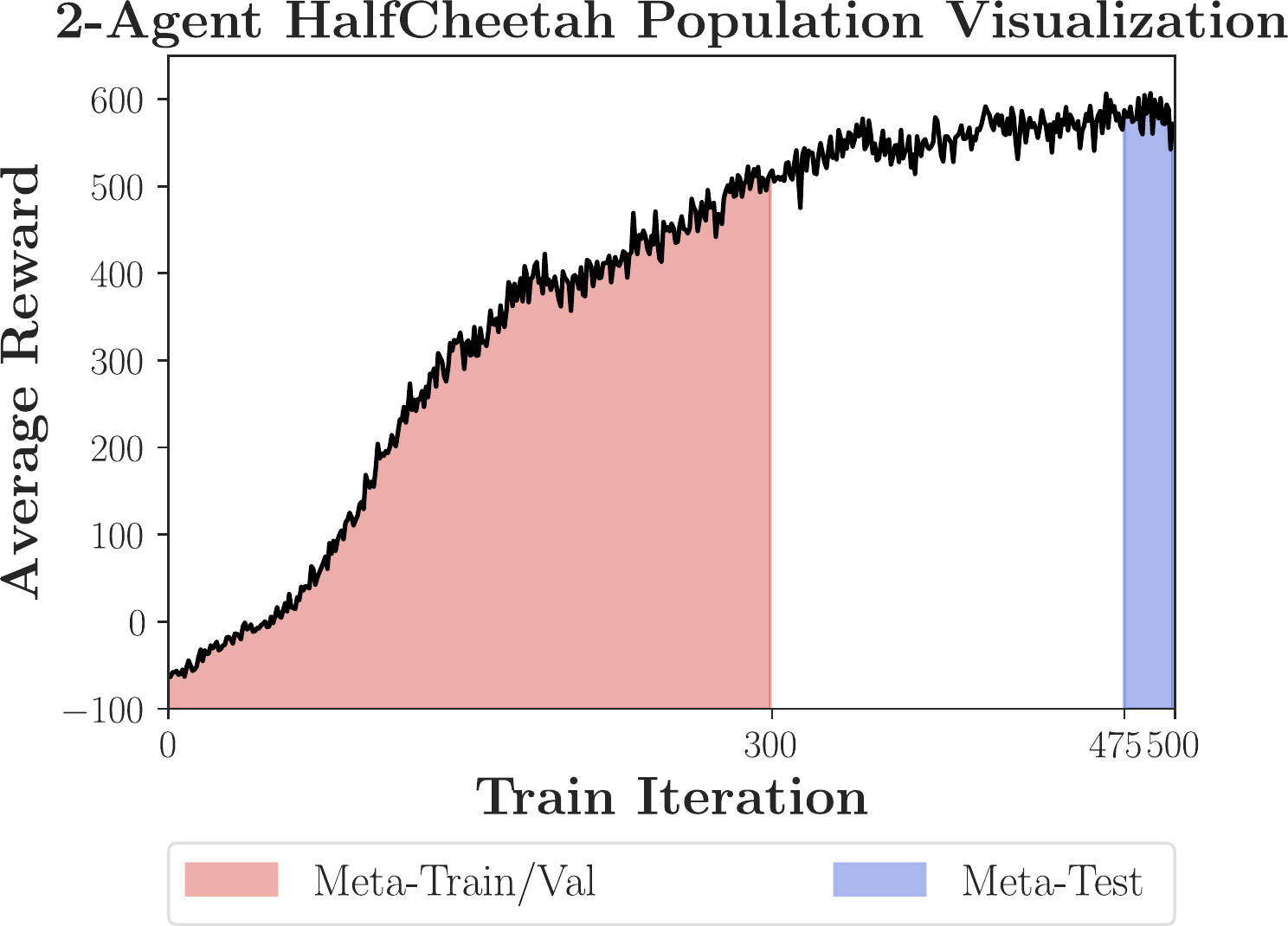}
  \caption{Visualization of a teammate $j$'s initial expertise in the $2$-Agent HalfCheetah domain, where the meta-test distribution has a sufficient difference to meta-train/val.}
  \label{fig:mujoco-teammate-vis}
\end{wrapfigure}
We used an open source implementation for multiagent-MuJoCo benchmark: \url{https://github.com/schroederdewitt/multiagent_mujoco}.
Agents in our experiments receive state observations that include information about all the joints.
For the meta-learning setup, we pre-train a teammate $j$ with an LSTM policy that has varying expertise in moving to the left direction. 
Specifically, we train the teammate up to $500$ train iterations and save a checkpoint at each iteration. 
Intuitively, as the number of train iteration increases, the teammate gains more expertise.
We then use the checkpoints from $0$ to $300$ iterations as the meta-train/val (randomly split them into $275$ for meta-training and $25$ for meta-validation) and from $475$ and $500$ iterations as the meta-test distribution (see~\cref{fig:mujoco-teammate-vis}). 
We construct the distribution with the gap to ensure that the meta-testing distribution has a sufficient difference to the meta-train/val so that we can test the generalization of our approach.
As in IPD and RPS, the teammate $j$ updates its policy based on the policy gradient with the linear feature baseline.

\clearpage
\section{Analysis on Joint Policy Dynamics}\label{sec:analysis-details}
\subsection{IPD}
\begin{figure}[H]
    \centering
    \includegraphics[width=0.59\linewidth]{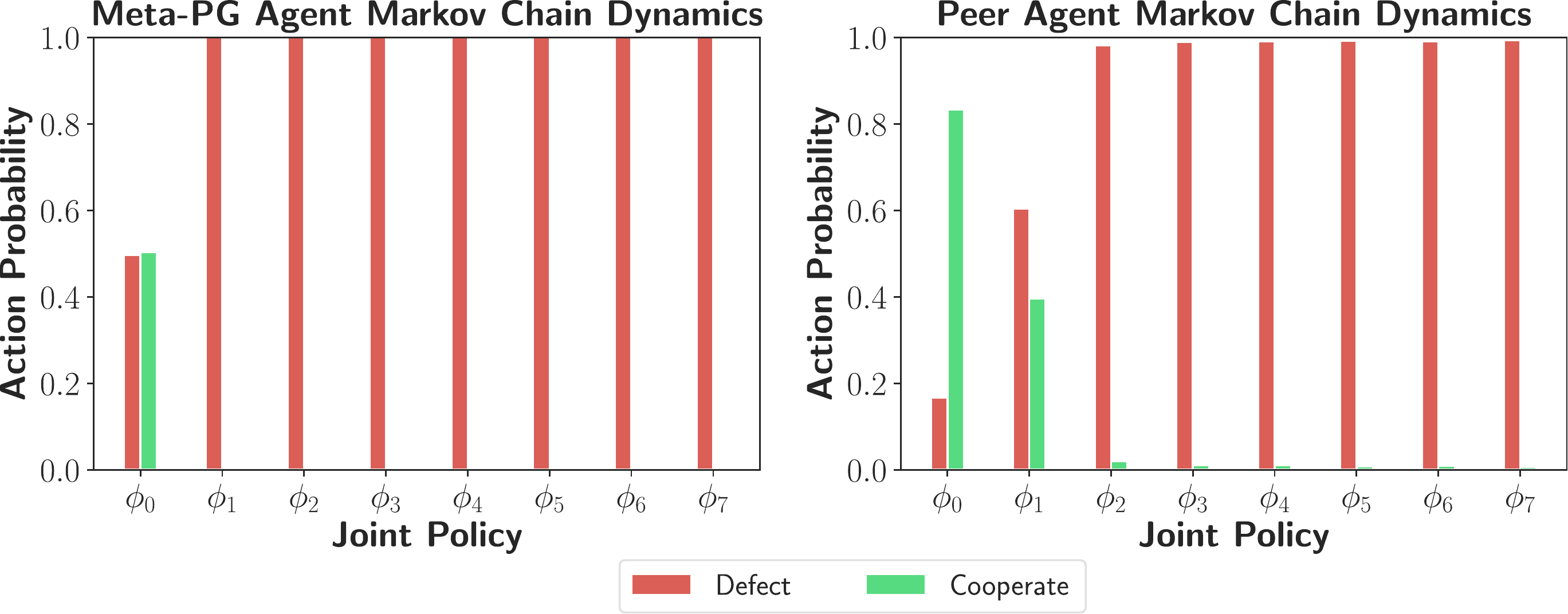}
    \caption{Action probability dynamics with Meta-PG in IPD with a cooperating persona peer}
    \label{fig:ipd-meta-pg-analysis}
\end{figure}
\begin{figure}[H]
    \centering
    \includegraphics[width=0.59\linewidth]{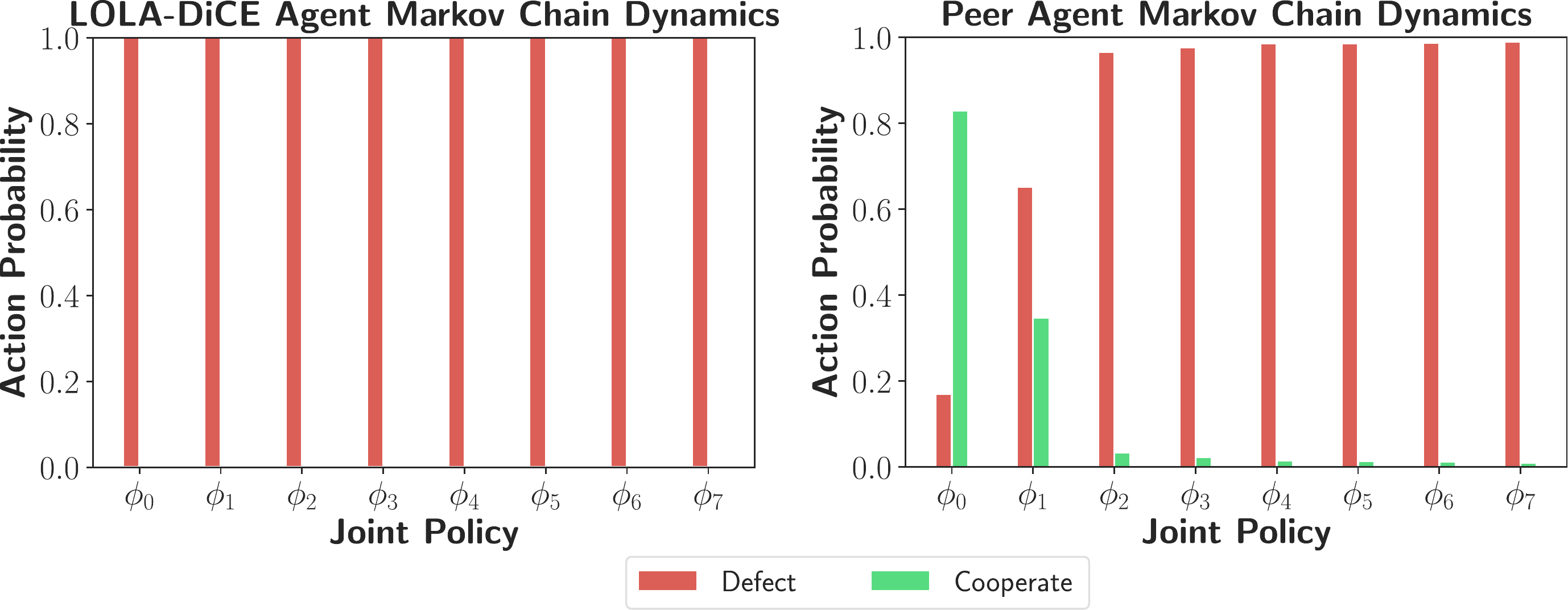}
    \caption{Action probability dynamics with LOLA-DiCE in IPD with a cooperating persona peer}
    \label{fig:ipd-dice-analysis}
\end{figure}
\begin{figure}[H]
    \centering
    \includegraphics[width=0.59\linewidth]{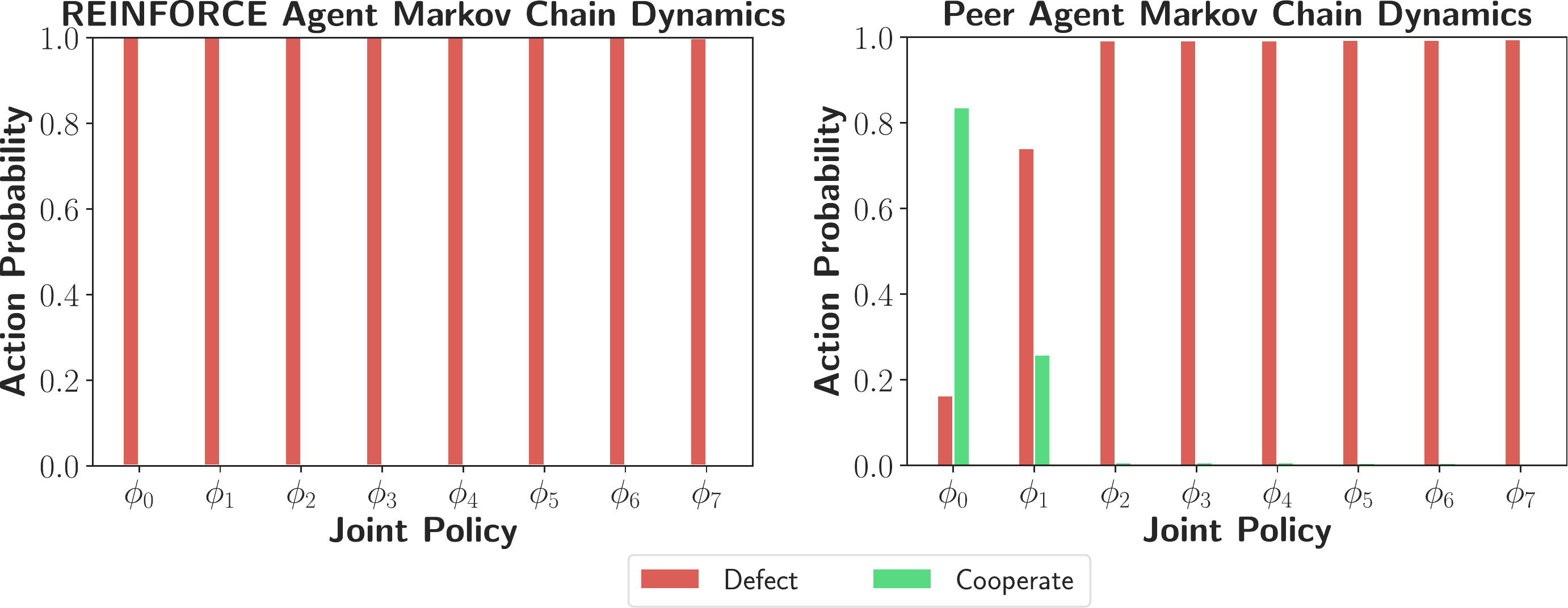}
    \caption{Action probability dynamics with REINFORCE in IPD with a cooperating persona peer}
    \label{fig:ipd-reinforce-analysis}
\end{figure}
\begin{figure}[H]
    \centering
    \includegraphics[width=0.59\linewidth]{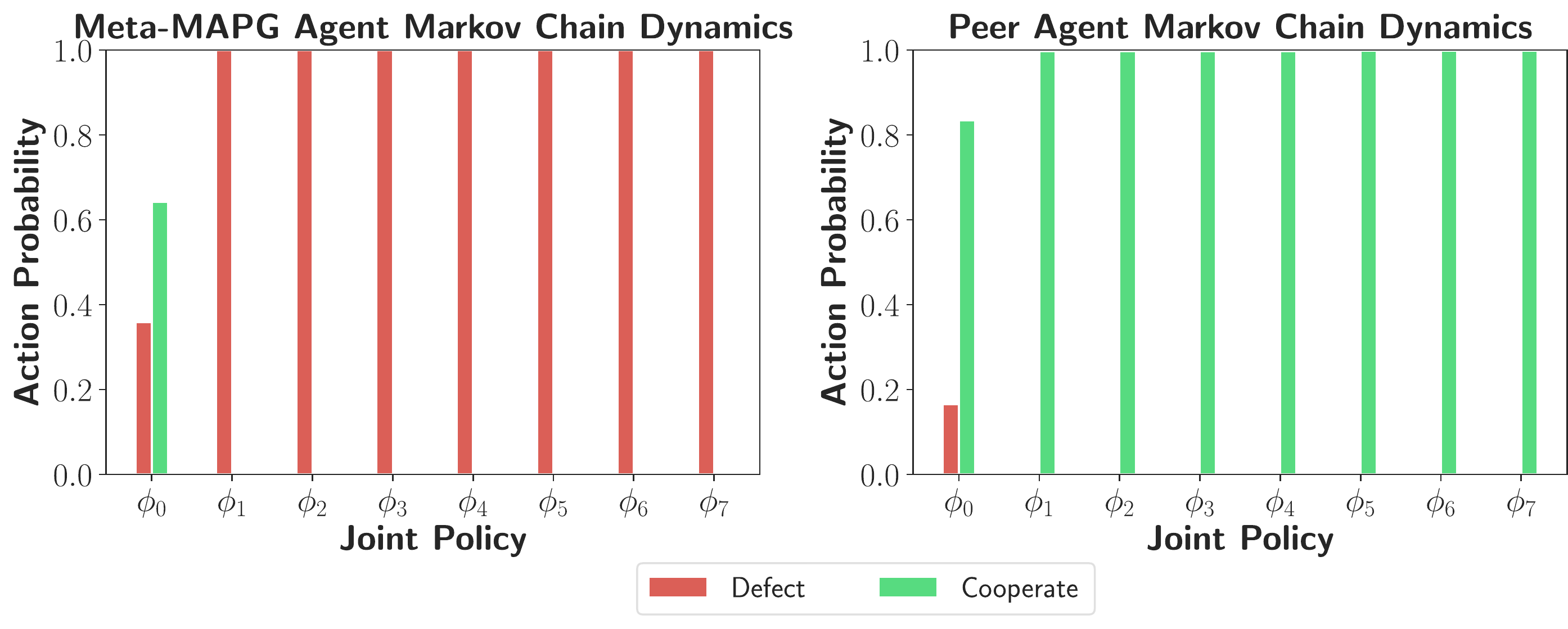}
    \caption{Action probability dynamics with Meta-MAPG in IPD with a cooperating persona peer}
    \label{fig:ipd-meta-mapg-analysis}
\end{figure}

\subsection{RPS}
\begin{figure}[H]
    \centering
    \includegraphics[width=0.59\linewidth]{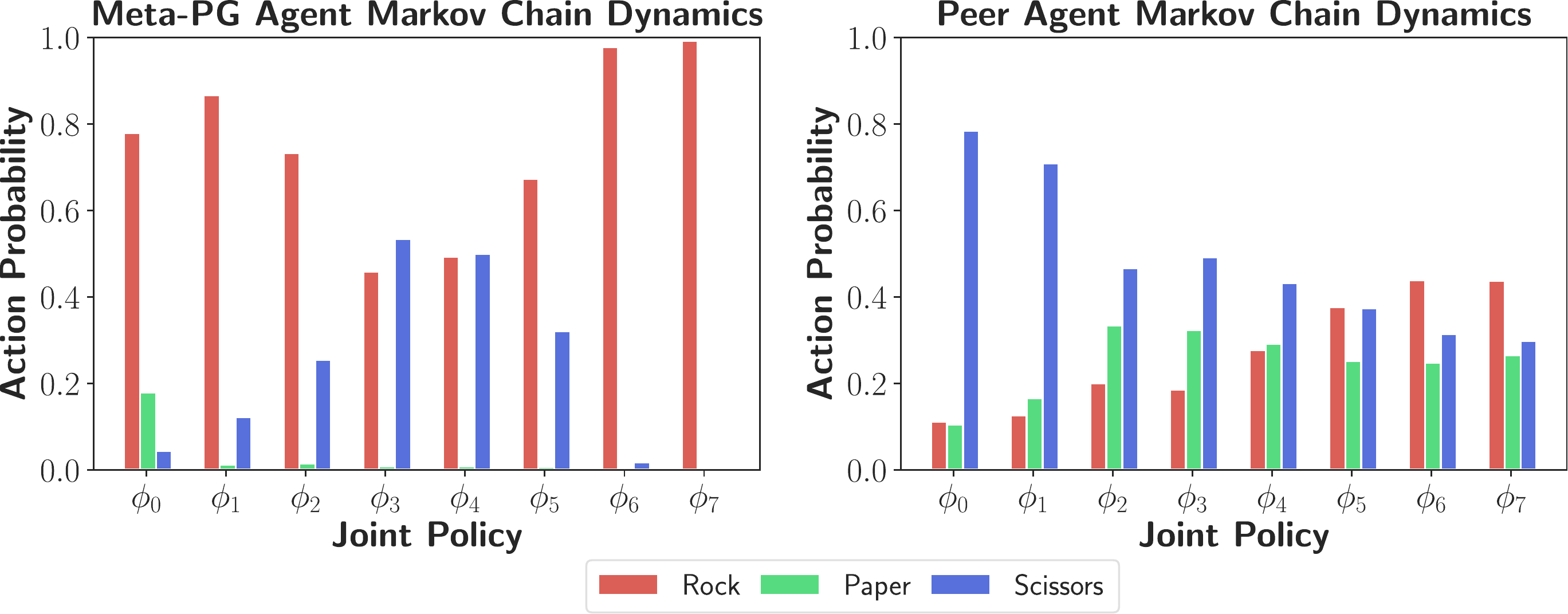}
    \caption{Action Probability Dynamics with Meta-PG in RPS with a scissors persona opponent}
    \label{fig:rps-meta-pg-analysis}
\end{figure}
\begin{figure}[H]
    \centering
    \includegraphics[width=0.59\linewidth]{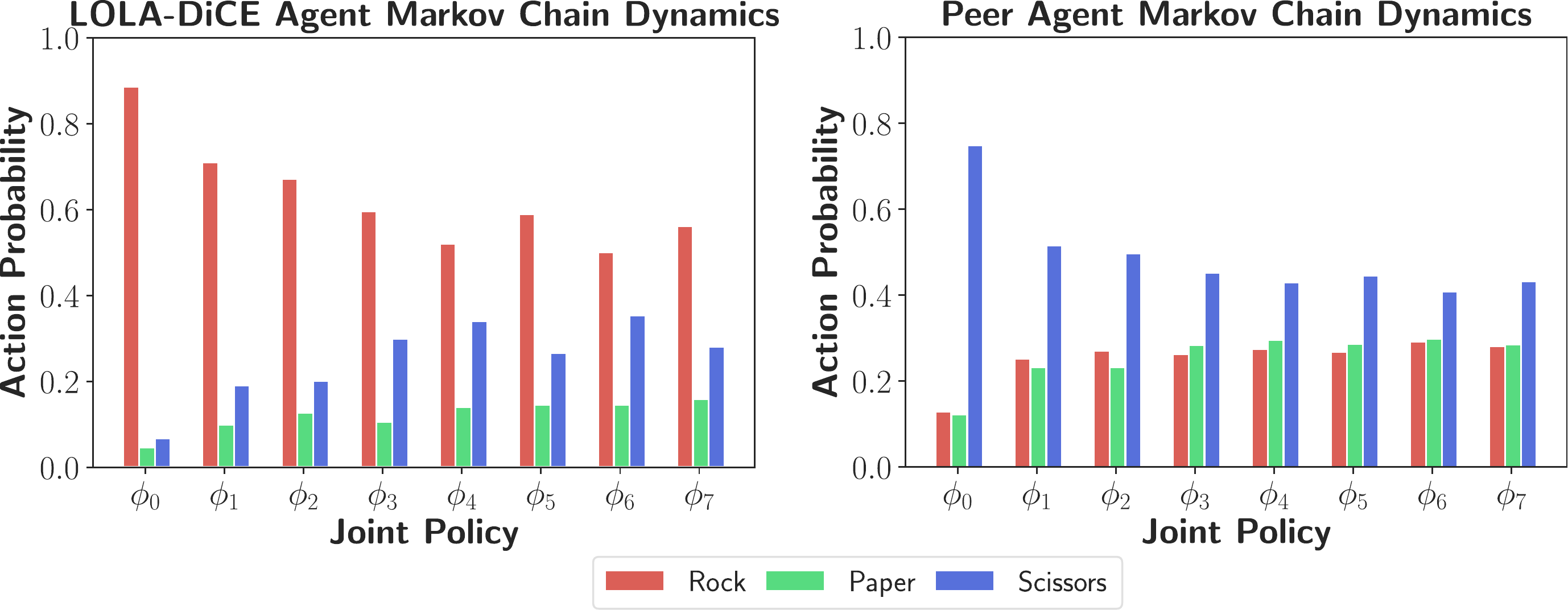}
    \caption{Action Probability Dynamics with LOLA-DiCE in RPS with a scissors persona opponent}
    \label{fig:rps-dice-analysis}
\end{figure}
\begin{figure}[H]
    \centering
    \includegraphics[width=0.59\linewidth]{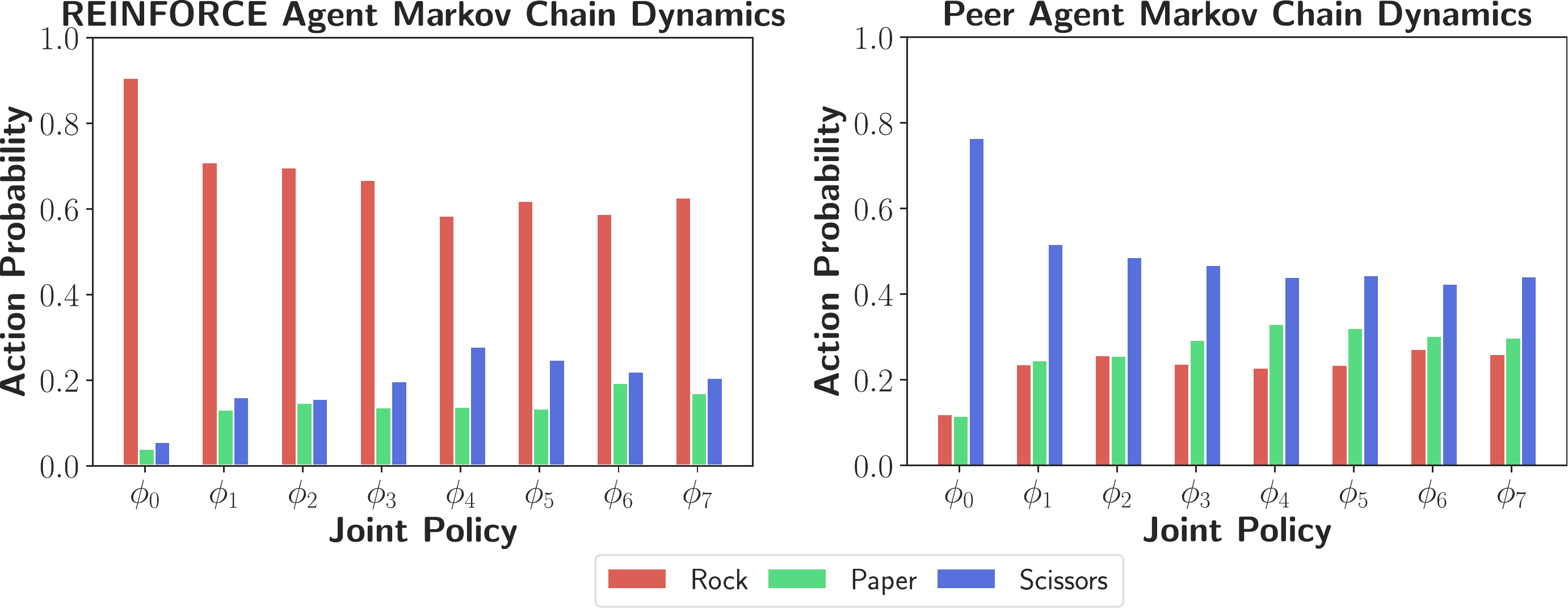}
    \caption{Action Probability Dynamics with REINFORCE in RPS with a scissors persona opponent}
    \label{fig:rps-reinforce-analysis}
\end{figure}
\begin{figure}[H]
    \centering
    \includegraphics[width=0.59\linewidth]{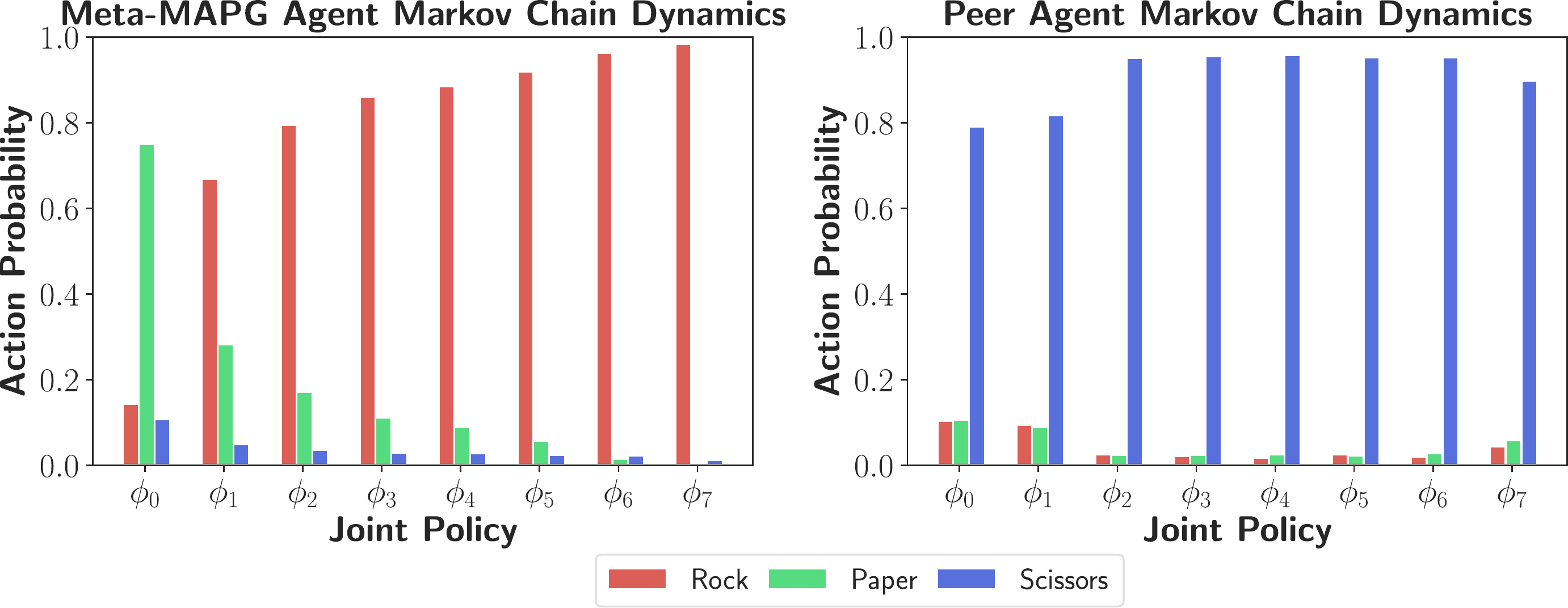}
    \caption{Action Probability Dynamics with Meta-MAPG in RPS with a scissors persona opponent}
    \label{fig:rps-meta-mapg-analysis}
\end{figure}

\section{Hyperparameter Details}\label{sec:hyperparameter-details}
We report our hyperparameter values that we used for each of the methods in our experiments:
\subsection{Meta-MAPG and Meta-PG}
\begin{table}[H]
\centering
\begin{tabular}{l|l}
Hyperparameter & Value \\ \hline
Trajectory batch size $K$ & 4, 8, 16, 32, 64 \\
Number of parallel threads & 5 \\
Actor learning rate (inner) & 1.0, 0.1 \\
Actor learning rate (outer) & 1e-4 \\
Critic learning rate (outer) & 1.5e-4 \\
Episode horizon $H$ & 150 \\
Max chain length $L$ & 7 \\
GAE $\lambda$ & 0.95 \\
Discount factor $\gamma$ & 0.96 \\
\end{tabular}
\caption{IPD}
\end{table}

\begin{table}[H]
\centering
\begin{tabular}{l|l}
Hyperparameter & Value \\ \hline
Trajectory batch size $K$ & 64 \\
Number of parallel threads & 5 \\
Actor learning rate (inner) & 0.01 \\
Actor learning rate (outer) & 1e-5 \\
Critic learning rate (outer) & 1.5e-5 \\
Episode horizon $H$ & 150 \\
Max chain length $L$ & 7 \\
GAE $\lambda$ & 0.95 \\
Discount factor $\gamma$ & 0.90 \\
\end{tabular}
\caption{RPS}
\end{table}

\begin{table}[H]
\centering
\begin{tabular}{l|l}
Hyperparameter & Value \\ \hline
Trajectory batch size $K$ & 64 \\
Number of parallel threads & 5 \\
Actor learning rate (inner) & 0.005 \\
Actor learning rate (outer) & 5e-5 \\
Critic learning rate (outer) & 5.5e-5 \\
Episode horizon $H$ & 200 \\
Max chain length $L$ & 2 \\
GAE $\lambda$ & 0.95 \\
Discount factor $\gamma$ & 0.95 \\
\end{tabular}
\caption{$2$-Agent HalfCheetah}
\end{table}

\subsection{LOLA-DiCE}
\begin{table}[H]
\centering
\begin{tabular}{l|l}
Hyperparameter & Value \\ \hline
Trajectory batch size $K$ & 4, 8, 16, 32, 64 \\
Actor learning rate & 1.0, 0.1 \\
Critic learning rate & 1.5e-3 \\
Episode horizon $H$ & 150 \\
Max chain length $L$ & 7 \\
Number of Look-Ahead & 1, 3, 5 \\
Discount factor $\gamma$ & 0.96 \\
\end{tabular}
\caption{IPD}
\end{table}

\begin{table}[H]
\centering
\begin{tabular}{l|l}
Hyperparameter & Value \\ \hline
Trajectory batch size $K$ & 64 \\
Actor learning rate & 0.01 \\
Critic learning rate & 1.5e-5 \\
Episode horizon $H$ & 150 \\
Max chain length $L$ & 7 \\
Number of Look-Ahead & 1 \\
Discount factor $\gamma$ & 0.90 \\
\end{tabular}
\caption{RPS}
\end{table}

\begin{table}[H]
\centering
\begin{tabular}{l|l}
Hyperparameter & Value \\ \hline
Trajectory batch size $K$ & 64 \\
Actor learning rate & 0.005 \\
Critic learning rate & 5.5e-5 \\
Episode horizon $H$ & 200 \\
Max chain length $L$ & 2 \\
Number of Look-Ahead & 1 \\
Discount factor $\gamma$ & 0.95 \\
\end{tabular}
\caption{$2$-Agent HalfCheetah}
\end{table}

\subsection{REINFORCE}
\begin{table}[H]
\centering
\begin{tabular}{l|l}
Hyperparameter & Value \\ \hline
Trajectory batch size $K$ & 4, 8, 16, 32, 64 \\
Actor learning rate & 1.0, 0.1 \\
Episode horizon $H$ & 150 \\
Max chain length $L$ & 7 \\
Discount factor $\gamma$ & 0.96 \\
\end{tabular}
\caption{IPD}
\end{table}

\begin{table}[H]
\centering
\begin{tabular}{l|l}
Hyperparameter & Value \\ \hline
Trajectory batch size $K$ & 64 \\
Actor learning rate & 0.01 \\
Episode horizon $H$ & 150 \\
Max chain length $L$ & 7 \\
Discount factor $\gamma$ & 0.90 \\
\end{tabular}
\caption{RPS}
\end{table}

\begin{table}[H]
\centering
\begin{tabular}{l|l}
Hyperparameter & Value \\ \hline
Trajectory batch size $K$ & 64 \\
Actor learning rate & 0.005 \\
Episode horizon $H$ & 200 \\
Max chain length $L$ & 2 \\
Discount factor $\gamma$ & 0.95 \\
\end{tabular}
\caption{$2$-Agent HalfCheetah}
\end{table}